\documentclass{article} %
\usepackage{collas2022_conference,times}
\usepackage[T1]{fontenc}

\usepackage{amsmath,amsfonts,bm}

\def\eqref#1{equation~\ref{#1}}

\def\1{\bm{1}}

\DeclareMathAlphabet{\mathsfit}{\encodingdefault}{\sfdefault}{m}{sl}
\SetMathAlphabet{\mathsfit}{bold}{\encodingdefault}{\sfdefault}{bx}{n}

\usepackage{hyperref}
\hypersetup{
    colorlinks=true,
    linkcolor=red,
    filecolor=magenta,      
    urlcolor=blue,
    citecolor=purple,
    pdftitle={ReactiveExploration},
    pdfpagemode=FullScreen,
    }

\usepackage{graphicx}
\usepackage{comment}
\usepackage{float}
\usepackage{xcolor}
\usepackage{amsmath}
\usepackage{amssymb}
\usepackage[abs]{overpic}
\usepackage[toc,page,header]{appendix}
\usepackage{minitoc}
\usepackage{algorithm}
\usepackage{algpseudocode}
\usepackage{subcaption} %
\usepackage[labelfont=bf]{caption}

\usepackage{array}
\usepackage{arydshln}
 \title{
Reactive Exploration to Cope with Non-Stationarity\\ in \Lifelong~Reinforcement Learning
}

\author{Christian Steinparz~\footnotemark[3]\ \ \thanks{Correspondence to: christian.steinparz@jku.at}
    \quad
    Thomas Schmied~\footnotemark[2]\quad 
    Fabian Paischer~\footnotemark[2] \quad 
    Marius-Constantin Dinu~\footnotemark[2] $^{\ }$\footnotemark[6] \quad  \\ 
    \bf Vihang Patil~\footnotemark[2] \quad
    Angela Bitto-Nemling~\footnotemark[2]$^{\ \ }$\footnotemark[4] \quad
    Hamid Eghbal-zadeh~\footnotemark[2] \quad
    Sepp Hochreiter~\footnotemark[2]$^{\
    \ }$\footnotemark[4] \\
    \\
  \footnotemark[2]\ \ ELLIS Unit Linz and LIT AI Lab, Institute for Machine Learning, Johannes Kepler University Linz, Austria \\  \footnotemark[3]\ \ Visual Data Science Lab, Institute of Compute Graphics, Johannes Kepler University Linz, Austria \\    
  \footnotemark[6]\ \ Dynatrace Research, Linz, Austria \\
  \footnotemark[4]\ \ Institute of Advanced Research in 
                    Artificial Intelligence (IARAI), Vienna, Austria\\
}

\newcommand\lifelong{lifelong}
\newcommand\Lifelong{Lifelong}
\usepackage{array}

\newcommand\Ba{\bm{a}}

\newcommand\Bo{\bm{o}}

 \newcommand{\dP}{\mathbb{P}} 
 \newcommand{\dR}{\mathbb{R}}

\newcommand{\cA}{\mathcal{A}} \newcommand{\cB}{\mathcal{B}}
 
\newcommand{\cE}{\mathcal{E}}

\newcommand{\cO}{\mathcal{O}} 
 \newcommand{\cR}{\mathcal{R}}
\newcommand{\cS}{\mathcal{S}} \newcommand{\cT}{\mathcal{T}}

\collasfinalcopy %

\begin{document}

\maketitle
\doparttoc
\faketableofcontents

\begin{abstract}

In \lifelong~learning, an agent learns throughout its entire life without resets, in a constantly changing environment, as we humans do.
Consequently, \lifelong~learning comes with a plethora of research problems such as continual 
domain shifts, which result in non-stationary rewards and environment dynamics.
These non-stationarities are difficult to detect and cope with due to their continuous nature.
Therefore, exploration strategies and learning methods are required that are capable of tracking the steady domain shifts, and adapting to them.
We propose \emph{Reactive Exploration} to track and react to continual domain shifts in \lifelong~reinforcement learning, and to update the policy correspondingly.
To this end, we conduct experiments in order to investigate different exploration strategies.
We empirically show that representatives of the policy-gradient family are better suited for \lifelong~learning, as they adapt more quickly to distribution shifts than Q-learning.
Thereby, policy-gradient methods profit the most from Reactive Exploration and
show good results in \lifelong~learning with continual domain shifts. Our code is available at: \url{https://github.com/ml-jku/reactive-exploration}.

\end{abstract}

\section{Introduction}
\label{sec:intro}

Existing Reinforcement Learning (RL) algorithms are tailored towards the limited setup of stationary environments. 
That is, considering a task in an episodic manner, with an assumption on stationary components of the underlying Markov Decision Process (MDP). 
This scenario does not reflect how we humans learn in the real world, which steadily evolves over time. 
Learning in non-stationary environments requires algorithms that are inherently capable of constantly dealing with domain shifts \citep{Widmer2004, adler2020cross}.
In the literature, several topics have been dedicated to study this problem.
For example, the paradigm of domain adaptation \citep{bendavid2010theory} comprises a single domain shift from a source to a target domain.
In the field of \lifelong~learning (or continual learning) \citep{thrun1995lifelong,thrun1998lifelong}, however, these domain shifts occur in a distinct continual fashion \citep{ring1994continual}.
While we humans easily adapt to such continual domain shifts in the real world, existing algorithms are often designed towards specific problem
settings, rendering them incapable of handling these continual shifts. 
In the context of \lifelong~learning, the horizon is unbounded, and agents must learn from potentially years of experiences under continual domain shifts.
The knowledge acquired by agents can then be used to facilitate transfer of the learned skills to various tasks to mitigate performance plateaus \citep{mitchellNeverEnding}.
An example of such problems are robots that constantly explore and ideally learn from previous experiences \citep{Woczyk2021Continual}. 

Many potential sources of non-stationarity exist in \lifelong~RL.
One way to induce non-stationarity is to change various components of the MDP over time, i.e. the reward function, or transition probabilities. 
Non-stationarity can also be introduced via altering observations or the set of actions over time \citep{ketharpal2020}.
These changes can occur gradually and over multiple steps \citep{besbes2014stochastic}, or abruptly. 
The latter is often referred to as piecewise stationary \citep{hartland2007change,garivier2008upperconfidence,cao2019nearly}.
Approaches that adapt to non-stationary environments have to cope with effects known as catastrophic forgetting or catastrophic interference, which is also referenced as the sensitivity-stability dilemma or stability-plasticity dilemma \citep{Hebb1949organization, Carpenter1987, McCloskey1989, mermillod2013stability, ehret2020continual}.
A policy learned under such conditions needs to retain valuable information from the past while still being able to quickly adapt to unseen situations.
Recent works in the context of \lifelong~RL also address this as forward transfer and backward transfer of knowledge \citep{wang2019forward}. 
Gaining new knowledge on current tasks may positively impact past beliefs of already learned tasks (backward transfer), or facilitate adaptation to future tasks (forward transfer).

In environments with domain shifts, it is difficult for a learning agent to keep track and react to changes.
The detection of such changes in continuous streams of data is often referred to as changepoint detection \citep{Kifer2004detecting,adams2007bayesian}.
Dealing with dynamic changepoints is crucial in coping with non-stationarity.
In this work, we propose to leverage Reactive Exploration to efficiently detect dynamic changepoints and facilitate recovery and forward transfer in \lifelong~learning.
Reactive Exploration predicts the expected future outcome using an observations model and a reward model, which enable detecting changes in the environment.
If the current input deviates strongly from the intrinsic belief of the agent, it is encouraged to explore, in order to quickly adapt to changes in the environment. 
Thus, Reactive Exploration assigns credit to exploring parts of the environment that diverge from its internal belief. 
We demonstrate that Reactive Exploration enables agents to deal with various types of non-stationarities.

In order to evaluate RL agents under these conditions, we systematically induce changes to the environment dynamics, observations, and the reward function.
Particularly, we compare Proximal Policy Optimization (PPO) \citep{schulman2017proximal} and Deep-Q Networks (DQN) \citep{mnih2013playing}, as the most popular representatives of on-policy and off-policy algorithms. 
First, we investigate the robustness of selected algorithms against steadily and abruptly changing reward functions.
We find that while PPO is able to recover in the face of gradually shifting rewards, DQN completely fails. 
Even for abrupt changes in the reward function, we observe that PPO is more successful at recovery. 
A thorough analysis of the different modules of DQN suggests that its shortcomings can be mostly attributed to (i) its lack of systematic exploration, and (ii) the replay buffer management. 
We find that injecting prior knowledge with respect to environmental changepoints into the exploration schedule and replay buffer allocation drastically improves stability of DQN.
However, such information is usually not available a-priori.
This renders DQN unsuitable for non-stationary problems.
Further, we induce non-stationarity in environment observations via different transformations, and find that even PPO completely fails in such a setup. 
Equipping the agent with Reactive Exploration significantly improves stability and robustness of PPO against the different forms of non-stationarity.
In summary, our in-depth analysis provides new insights on the problem of \lifelong~RL, which result in the following contributions:
\begin{enumerate}
\item PPO is more robust to domain shifts over time, compared to DQN.
\item DQN is not suitable for non-stationarity when no additional information is given on occurring shifts.
\item For severe domain shifts, both DQN and PPO fail to adapt.
\item Reactive Exploration helps PPO to recover, even for severe shifts.
\end{enumerate}

\section{Related Work}
\label{sec:related_work}

Learning several tasks sequentially is necessary to tackle \lifelong~learning problems.
Agents need to retain already gained knowledge, but also remain pliable to acquire new knowledge.
Ideally, this is achieved by leveraging past experiences to effectively build up knowledge representations that are useful across multiple tasks, allowing for forward and backward transfer of knowledge.
Early neural network architectures were shown to be unsuitable for continual domain shifts, and incapable of coping with effects of catastrophic forgetting or the stability-plasticity dilemma~\citep{french1999catastrophic}.
The field of \lifelong~learning has emerged to study general approaches to alleviate these issues.

Some approaches in \lifelong~learning address continual domain shifts via rehearsal or pseudo-rehearsal methods \citep{Parisi2018lifelong, Hayes2019memory}, the reduction of representational overlap \citep{Serra2018overcoming, Wortsman2020supermasks} or architectural strategies \citep{Rusu2016ProgressiveNN, Fernando2017PathNetEC, Mallya2018packnet, Xu2018reinforced, Maltoni2019continuous}, regularization strategies for consolidating parameters \citep{kirkpatrick2017overcoming, zenke2017continual}, and correcting for domain shifts  \citep{Li2017learning, Lee2017overcoming, Zeno2021TaskAgnosticCL}.
Other approaches consider meta-learning \citep{Schmidhuber1987, Bengio1992}, which uses task-specific inner loops and task agnostic outer loops to learn supportive representations that are helpful across tasks \citep{Javed2019Meta, Gupta2020maml}.
\Lifelong~learning can also be addressed via multi-task learning \citep{caruana1997multitask}, by learning joint embedding spaces over multiple tasks at the same time \citep{mitchellNeverEnding}.

In the context of non-stationarity and \lifelong~RL specifically, continual domain shifts may occur in state or observations spaces, action spaces, transition dynamics, reward functions, or combinations thereof \citep{cheung2020, igl2020, ketharpal2020}.
Commonly, non-stationarity is encountered in transition dynamics or reward functions.
To cope with such non-stationarities an agent must be equipped with a mechanism that is capable of detecting environmental changes to facilitate its adaptation to a new domain.
\citet{padakandla2019reinforcement} combine Q-learning \citep{watkins1992} with a changepoint detection algorithm \citep{prabuchandran2022changepoint} to handle changing environment dynamics.
\citet{canonaco2020model} combine policy gradient algorithms with the CUSUM \citep{poor1996detection} approach for changepoint detection.
\citep{alegre2021minimum} utilizes an ensemble of learned dynamics models to compute change statistics and deploy different policies for different contexts.
Instead of detecting changepoints, \citet{flennerhag2022bootstrapped} meta-learn an exploration schedule to cope with a non-stationary reward function.
In contrast, Reactive Exploration is a mechanism that allows for both, detecting changepoints while simultaneously assigning credit to exploration of novel states that have undergone change.

In other settings, such as multi-armed bandit (MAB) problems, \citet{raj2017taming} take a Bayesian approach and investigate Thompson sampling and its relation to non-stationarity, \citet{wu2018learning} propose a contextual bandit \citep{lattimore2020bandit} algorithm to detect changes based on its reward estimation confidence, and update the bandit arm selection strategy respectively.
\citet{besbes2014stochastic} characterize the regret in MAB problems with non-stationary rewards such that there is a link between reward variation and minimal achievable regret to factor in a variation budget to allow for temporal changes. 
\citet{zhao2020simple} propose a method to overcome non-stationarity by investigating a periodically restarted Upper-Confidence-Bound algorithm, while \citet{russac2019weighted} propose an algorithm based on discounted linear regression.
\citet{hong2021} focus on abrupt changes in contextual bandits, and develop an off-policy method to learn sets of policies corresponding to a categorical latent state for individual phases of non-stationarity.
\citet{besbes2019optimal} introduce a non-parametric statistical model to track changes in the reward function over time.
However, MAB settings do not consider state-transition dynamics. 

Additional problems in \lifelong~RL include credit assignment \citep{sutton1984temporal, arjona2019rudder, patil2020align, holzleitner_convergence_2020,
widrich_modern_2021,Dinu2022}, abstraction \citep{mitchell2022abstraction,paischer2022helm}, and exploration \citep{Sutton1998}. 
Several works address the exploration problem by novelty-based \citep{burda2019exploration},
count-based, \citep{strehl2008analysis,bellemare2016unifying,ostrovski2017count,martin2017count,tang2017study,machado2020count},
curiosity-based \citep{Schmidhuber1991possibility,schmidhuber1990making}, and model-based \citep{stadie2015incentivizing,pathak2017curiosity,achiam2017surprise} approaches.
Additionally, \citet{oudeyer2007what} use intrinsic motivation, and \citet{raileanu2020ride} include novelty measures for exploration in procedurally generated environments.
In the context of hierarchical RL, \citet{mcgovern2001automatic,goel2003subgoal} tackle exploration for temporally extended actions. 
\citet{garcia2019meta} address the problem of optimizing the exploration strategy in a meta-MDP.
Other strategies include memory-based exploration strategies such as episodic memory \citep{badia2020never} or planning with worst-base planning evolution \citep{Lecarpentier2019NonStationaryMD}, Go-Explore \citep{ecoffet2019go}, and optimal experiment design \citep{Cohn1994Advances,Storck1995}.
Nonetheless, the exploration/exploitation trade-off has been studied little in the context of non-stationary environments.
Current state-of-the-art exploration mechanisms such as RIDE \citep{raileanu2020ride} or NovelD \citep{zhang2021noveld} rely on episodic counts and are thus not applicable to the infinite horizon setting out-of-the-box. 
Methods such as Intrinsic Curiosity Module (ICM)~\citep{pathak2017curiosity} and Random Network Distillation (RND)~\citep{burda2019exploration} do not have this limitation.

\begin{figure}[h!]
    \centering
    \includegraphics[width=1.0\linewidth]{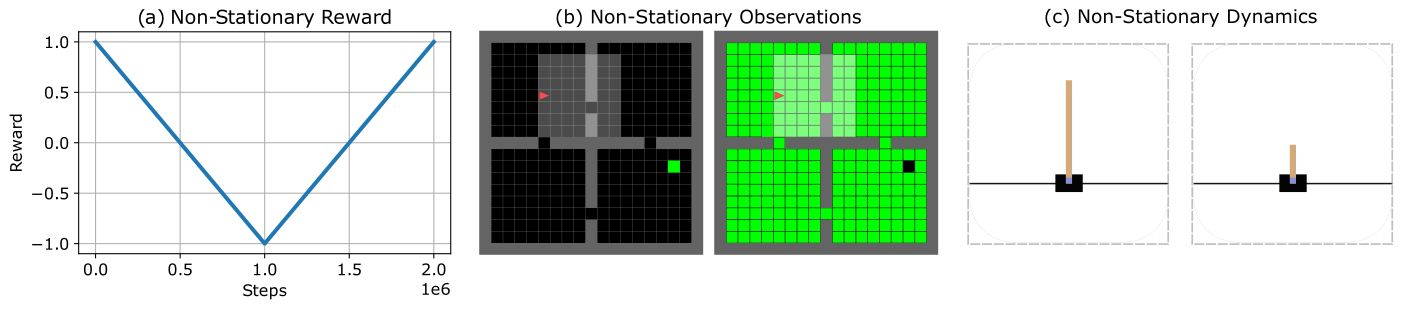}
    \caption{Different types of Non-Stationarity. (a) changing reward function over time; (b) shift in the observation distribution of a gridworld \citep{gym_minigrid}; (c) distribution shift in the dynamics of the CartPole balancing system \citep{openai}, altering the length of the pole results in different dynamics.}
    \label{fig:types_of_non-stationarity}
\end{figure}

\section{Problem Statement}
\label{sec:problem_statement}

One of the big problems in \lifelong~RL is the capability to adapt to new distributions.
Agents that are not equipped with tools to deal with domain shifts drastically fail to solve a given task in \lifelong\ RL.
The problem of dealing with domain shifts can occur in different aspects, and in varying levels of severity and intervals.
Some of the most important ones which are at the core of \lifelong\ RL are: (a) changes in the rewards, (b) changes in the observations received from the environment, and (c) changes in the environment dynamics.
Each of these introduce a new kind of challenge to the agent. 
In (a) for example, an agent must discover the new goal to be accomplished and re-evaluate its behavior accordingly, as the reward distribution shifts (Figure \ref{fig:types_of_non-stationarity}, left). 
In (b) the agent must generalize to the observation distribution that presents a new representation of the world (Figure \ref{fig:types_of_non-stationarity}, middle), while in (c) the agent must approximate changes occurring in the underlying physics (Figure \ref{fig:types_of_non-stationarity}, right).
All these options are unique in nature and present a set of challenges that might differ from one another.

We as humans deal with similar situations on a daily basis, and successfully adapt our strategies to the evolving world around us.
When facing any of the aforementioned problems, we first perceive the occurrence of change, and then accustom to those changes as we realize the previous strategy no longer works.
In other words, we detect the changes, and react accordingly.
A necessary component facilitating this capability is our internal model of the world, which stores all necessary information to reliably predict future outcomes \citep{Forrester1971}.
This internal model allows us to act upon internal predictions of how the world will behave in the future \citep{Maus2013MotionDependentRO}.
In turn, we are inherently capable of detecting changepoints when our perception of the world strongly deviates from our internal expectation of how the world will behave in the future.
Similarly, we attempt to bridge the gap for changepoint detection and adaptation in the realm of \lifelong\ RL.
In this regard, we define Reactive Exploration as a mechanism that intrinsically motivates the agent to explore after a changepoint has been detected.
Reactive Exploration equips an agent with the means to detect changes in the distribution of dynamics, observations, or rewards. 
Furthermore, it forces the agent to explore regions of the state space that have undergone change, immediately after they experienced change.

In the following, we consider the setting of a partially observable Markov Decision Process (POMDP,~\citealp{kaelbling_planning_1998}), which is defined as a 7-tuple ($\cS,\cA,\cT,\cR,\Omega,\cO$,$\gamma$), where $\cS$ is the set of states, $\cA$ is a set of actions, $\cT: \cS \times \cA \mapsto \dP(\cS)$ is the transition function which takes a state $s \in \cS$ and an action $a \in \cA$ and maps it to a probability distribution over states $\cS$.
Further, $\cR: \cS \times \cA \mapsto \dR$ is the reward function which takes a state and an action and maps it to a scalar reward, $\Omega$ is the set of observations, $\cO: \cS \times \cA \mapsto \dP(\Omega)$ is the observation function that yields a distribution over observations given a state-action pair, and $\gamma \in [0,1)$ is the discount factor. 
The policy $\pi_\theta: \Omega \mapsto \cA$ is parameterized by $\theta$ and maps an observation $\Bo \in \cO$ to an action $a \in \cA$.
Moreover, we consider the infinite horizon case, in which the episode horizon is unbounded. 
In Section~\ref{solution-proposal}, we present our Reactive Exploration method, in order to cope with non-stationarities and continual domain shifts in \lifelong~RL.
We show empirical results in Section~\ref{sec:experimental_results} which illustrate that state-of-the-art RL algorithms struggle to solve tasks under different non-stationarities, and how Reactive Exploration can enable the agents to cope with these issues.

\begin{figure*}[ht]
    \centering
    \begin{subfigure}[t]{0.4\textwidth}
         \centering
         \raisebox{0.25\height}{\includegraphics[width=.6\linewidth]{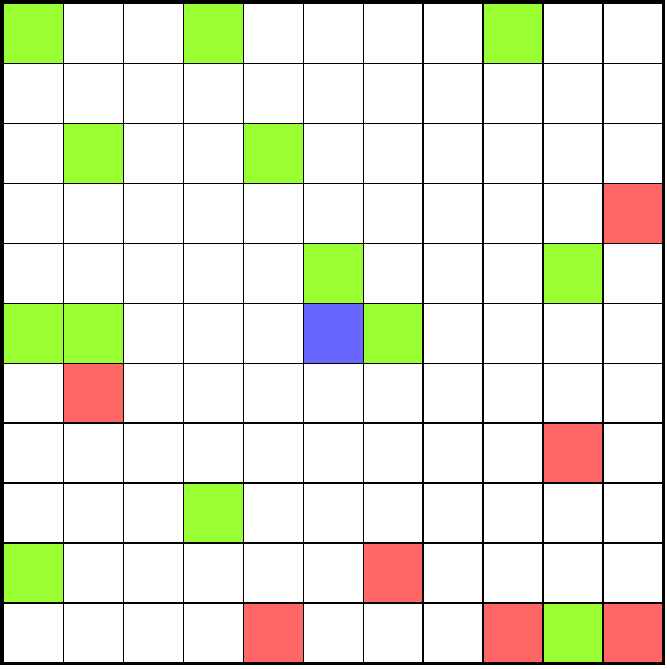}}
         \caption{}
     \end{subfigure}%
     ~
     \begin{subfigure}[t]{0.6\textwidth}
         \centering
         \includegraphics[width=1.\linewidth]{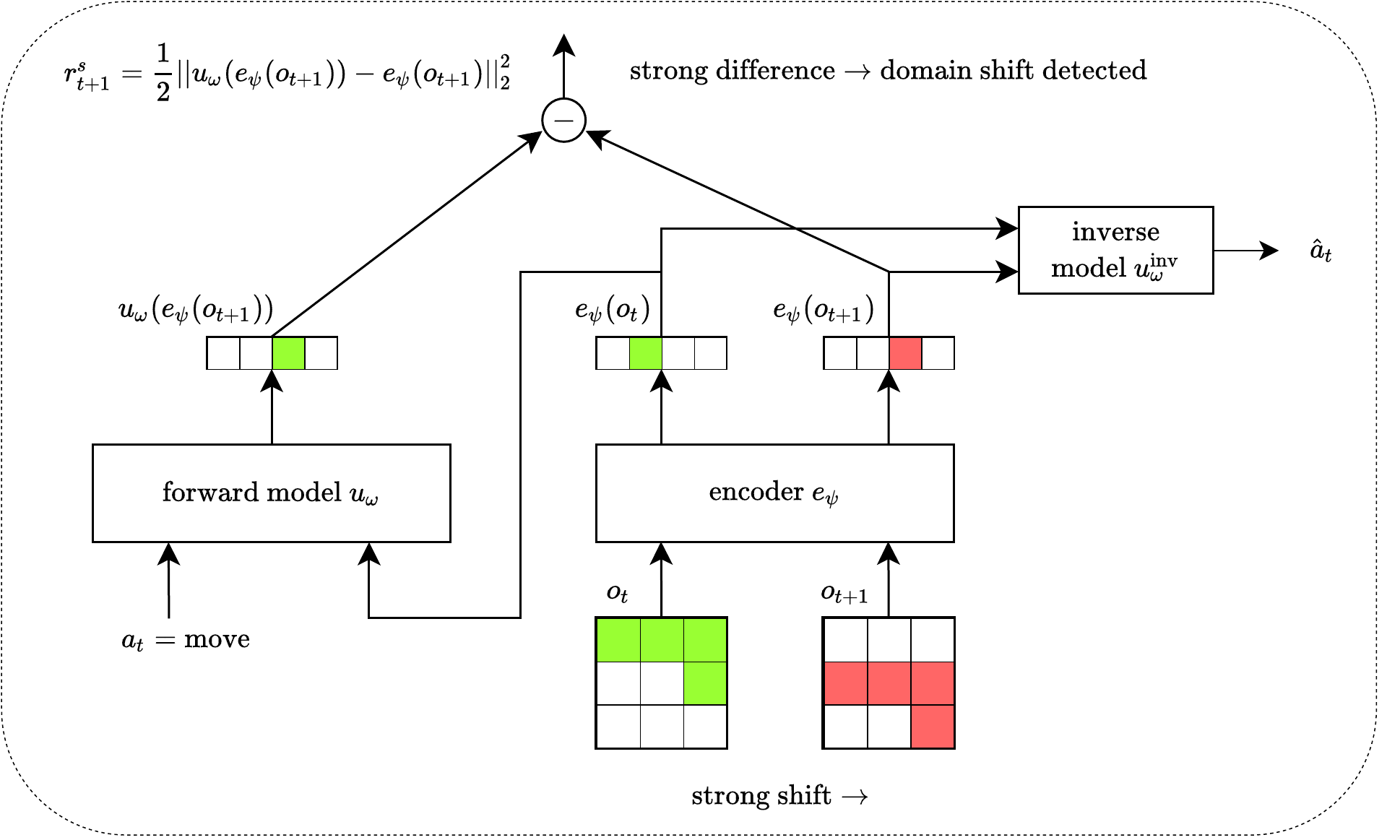}
         \caption{}
     \end{subfigure}
    \caption[]{(a) Layout of the JellyBean-Base environment: image observations are agent-centered, walking over items adds them to the agents' inventory and yields reward. (b) Intrinsic Curiosity Module as one specific instance of Reactive Exploration detecting a distribution shift in the observations as colors are changing. We discuss and apply alternative instantiations in Appendix~\ref{appendixsub:other-exploration-methods}.}
    \label{fig:jbw-v2}
\end{figure*}

\section{Reactive Exploration as Remedy for Non-Stationarity}
\label{solution-proposal}

Exploration strategies can help to track non-stationarities and to cope with their drawbacks. 
After encountering a changepoint, what an agent has come to believe about certain aspects of the environment may become incorrect.
We are interested in the exact point of change and the nature of change. 
If these were known, the agent could simply resample transitions of the affected parts of the environment. 
As a result, the agent could then correct its policy accordingly.
However, this is usually not the case since agents do not receive any information about changes in the environment.
By equipping an agent with an exploration mechanism, it can be encouraged to re-explore parts of the environment that changed over time.
Reactive Exploration is a mechanism that enables the agent to react to the novelties observed in the environment after a domain shift occurred. 
More specifically, in Reactive Exploration, a mechanism is used to detect changes in the environment and controls the amount of exploration directed towards changed regions, via an intrinsic reward signal.
We illustrate the effectiveness of Reactive Exploration and its capability to track changes in the environment and encourage exploration towards changed regions.

Our Reactive Exploration is based on the Intrinsic Curiosity Module \citep{pathak2017curiosity}, which consists of a forward and an inverse dynamics model.
Both models share an encoder $e_\psi: \Omega \mapsto \dR^d$ that maps an observation $o_t \in \Omega$ to a vector of dimension $d$.
The forward model $u_\omega: \dR^d \times \cA \mapsto \dR^d$ maps the encoded observation of the current timestep $e_\psi(\Bo_t) \in \dR^d$ and an action $a_t \in \cA$ to the encoded next timestep $e_\psi(\Bo_{t+1}) \in \dR^d$.
The inverse model $u^{inv}_\omega: \dR^d \times \dR^d \mapsto \dP(\cA)$ takes as input the encoded current observation $e_\psi(\Bo_t) \in \dR^d$ and the encoded next observation $e_\psi(\Bo_{t+1}) \in \dR^d$ and predicts a probability distribution over actions $\dP(\cA)$.
Furthermore, we add a reward model $g_\nu: \Omega \times \cA \mapsto \dR$ that approximates the extrinsic reward $r_{t+1}^e$, given the current observation $\Bo_t \in \Omega$ and action $a_t \in \cA$.
The training of the dynamics models is decoupled from training the policy $\pi_\theta$ and can happen on different timescales.
In the off-policy case, the entire replay buffer is used for training the dynamics models which contains data collected by the behavioral policy $\pi_\beta$ while in the on-policy case the data stems from the current policy $\pi_\theta$.

The intrinsic reward is defined as the prediction error of the observation prediction model and the reward model.
Since these error terms appear in different scales and are in principle unbounded we introduce three weighting factors $\alpha \in [0,1]$, $\beta \in [0,1]$, and $\lambda \in [0,1]$ with the constraint $\lambda = 1 - \alpha - \beta$ to properly combine the extrinsic with the intrinsic rewards.
In principle, any novelty-based exploration mechanism may be used.
Depending on the task at hand different components may be neglected, i.e. when dealing with shifting dynamics, the reward model does not provide additional useful information.
Algorithm~\ref{alg:reactive_exp} outlines the policy update with added Reactive Exploration.

\begin{figure}[!ht]
\begin{algorithm}[H]
\caption{Policy Update with Reactive Exploration}
\begin{algorithmic}
	\Require Weighting factors $\alpha \in [0, 1]$, $\beta \in [0, 1]$ and $\lambda \in [0, 1]$, policy model $\pi_\theta$, observation model $u_\omega$, reward model $g_\nu$, parametrized by $\theta, \omega, \nu$ respectively, Buffer $\cB$, and environment $\cE$
	\Ensure $\lambda = 1 - \alpha - \beta$
	\State \text{Initialize} $\theta, \omega, \nu$, $\cB \leftarrow \emptyset$
	\Comment{Initialize parameters and buffer}
    \Repeat
        \For{$k$ iterations}
    	\State $(\Bo_{t}, \Ba_{t}, r^{e}_{t+1}, \Bo_{t+1}) \sim \cE \text{~with action~} \Ba_{t} \sim \pi_\theta(\cdot \mid \Bo_t)$ 
    	\Comment{Sample transition from environment}
    	\State $\cB \leftarrow \cB \cup \{ \Bo_t, \Ba_t, \Bo_{t+1}, r^{e}_{t+1} \}$
    	\Comment{Add sample to buffer}
    	\EndFor
    	\For{$j$ update steps}
    	\State $(\Bo_t, \Ba_t, \Bo_{t+1}, r^{e}_{t+1}) \sim \cB$
    	\Comment{Sample transitions from buffer}
    	\State $r^s_{t+1} = u_\omega(\Bo_{t}, \Ba_{t}, \Bo_{t+1})$
    	\Comment{Compute intrinsic reward with observation model}
    	\State $\hat{r}^{e}_{t+1} = g_\nu(\Bo_{t}, \Ba_{t})$
    	\Comment{Predict extrinsic reward with reward model}
    	\State $r^{\text{r}}_{t+1} = 1 / 2 (\hat{r}^{e}_{t+1} - r^{e}_{t+1})^2$
    	\Comment{Compute intrinsic reward of the reward model}
    	\State $r_{t+1} = \alpha r^s_{t+1} + \beta r^r_{t+1} + \lambda r^e_{t+1}$
    	\Comment{Compute combined reward}
    	\State $\theta \leftarrow \texttt{UpdatePolicy}(\pi_\theta, \Bo_t, \Ba_t, \Bo_{t+1}, r_{t+1})$
    	\Comment{Update policy}
    	\State $\nu \leftarrow \texttt{UpdateRewardModel}(g_\nu, \Bo_t, \Ba_t, r^{e}_{t+1})$
    	\Comment{Update reward model}
    	\State $\omega \leftarrow \texttt{UpdateObservationModel}(u_\omega, \Bo_t, \Ba_t, \Bo_{t+1})$
    	\Comment{Update observation model}
    	\EndFor
    \Until done
\end{algorithmic}
\label{alg:reactive_exp}
\end{algorithm}
\setcounter{algorithm}{0}
\vspace{-1.5em}
\captionof{algorithm}{Reactive exploration for detecting non-stationarities and adjust to them. The observation model $u_\omega$ can be instantiated with any novelty-based exploration module. The reward model $g_\nu$ may be neglected if no shift in the reward function occurs. In the case of on-policy RL the buffer $\cB$ is instantiated as rollout buffer, while in the off-policy case it is instantiated as replay buffer \citep{lin_self-improving_1992}.}
\vspace{-1.5em}
\label{fig:algo1}
\end{figure}

\medskip

We choose this particular setup as Reactive Exploration mechanism since it provides several benefits when dealing with non-stationarities.
First, the forward dynamics model learns to predict the next observation and its prediction error reflects when changes in the environment dynamics occur.
After such a shift, it encourages the agent to explore regions of the observation space that yield a high error.
Second, the inverse dynamics model partially alleviates the so-called ``noisy-TV'' problem \citep{burda2019exploration} apparent in stochastic environments, which refers to an endless stream of exploration-based rewards being generated by randomly changing aspects of an environment.
The inverse dynamics model forces the shared encoder to neglect features that are created randomly and are therefore unpredictable.
Thereby, the prediction of the forward dynamics model becomes invariant to random effects in the environment.
Moreover, our setup does not rely on episodic information to compute the intrinsic rewards, as compared to other state-of-the-art exploration algorithms \citep{raileanu2020ride,zhang2021noveld}.
Finally, the reward model allows detecting changes in the reward function.

\section{Empirical Analysis of Reactive Exploration}
\label{sec:experimental_results}

We provide compelling evidence which supports the use of Reactive Exploration to cope with domain shifts.
In Section~\ref{subsec:environment} we elaborate on the environment used for our setup which allows introducing various non-stationarities as discussed in Section \ref{sec:problem_statement} in a controlled manner.
We train two well-established algorithms (PPO and DQN) in our environment and present empirical evidence in three parts:
First, in Section~\ref{subsec:ppo_vs_dqn_nonstationarity}, 
we show which types of non-stationarity are especially difficult for agents to cope with by altering the reward function. 
We find that PPO consistently outperforms DQN when confronted with various domain shifts. 
Our analysis of the shortcomings of DQN identifies two major reasons for its plummeting performance (Section \ref{sec:dqn_failures}).
On one hand, the replay buffer management is critical to avoid interference of transitions collected under different non-stationarities.
On the other hand, a more informed exploration strategy proves to be extremely useful in the face of non-stationarity.
In Section~\ref{subsec:breaking_ppo} we apply transformations to enforce non-stationary observations which are severe enough to break the PPO agents.
Following this, in Section~\ref{subsec:fixing_ppo}, we add Reactive Exploration to PPO, which drastically improves performance under domain shifts.
These results demonstrate the usefulness of Reactive Exploration, which helps PPO to recover and can facilitate forward transfer between stationarities.
Details on hyperparameters, as well as the model architectures, can be found in  Appendix~\ref{appendix:experiment-setup}.

\subsection{Environments}
\label{subsec:environment}

We use Jelly Bean World (JBW) \citep{platanios2020jelly}, to design environments with different levels and kinds of non-stationarities.
JBW is commonly used in evaluating agents in \lifelong~learning, and supports infinite horizon RL, non-stationarity, multi-task, and multi-modal learning \citep{ngiam2011multimodal}.
Environments in JBW are procedurally generated POMDPs, where only a fraction of the created world is visible to the agent.
By default, JBW provides multi-modal observations comprising images, as well as a vector representing the simulated diffusion of scent.
It allows simple and efficient control over task design, and enables incorporating various types of non-stationarities.
In our experiments, we configure the environment to contain two easily reachable types of items. 
Particularly, the items are represented by green and red pixels, which randomly occur at the same frequency. 
In the stationary setting, the collection of a green item yields +1.0 reward, while collecting a red item yields -1.0 reward. Collected items are added to the inventory of the agent.
We refer to the stationary environment with fixed reward function as JellyBean-Base.
A sketch of the JellyBean-Base environment is depicted in Figure~\ref{fig:jbw-v2}. 

Different types of non-stationarities are incorporated into JellyBean-Base by altering the reward function, and observations.
Similar to \citet{kessler2022same}, we introduce a distribution shift in the reward function to generate interfering tasks.
To this end, we consider two different properties of non-stationarity: (i) the smoothness of change, and (ii) the interval of change.
The former defines whether changes occur gradually over a certain timespan, or abruptly, i.e. piecewise.
The interval is defined as the period after which the reward function is fully inverted. 
In this regard, we design four different non-stationary reward functions: 1) abrupt inversions every $n=1\text{e}6$ steps, 2) gradual inversions over $n=1\text{e}6$ steps, 3) abrupt inversions every $n=1\text{e}5$ steps, and 4) gradual inversions over $n=1\text{e}5$ steps. 
All of these inversions occur repeatedly throughout an experiment over $2\text{e}6$ interaction steps with the environment.

Further, we incorporate two different cases of non-stationarities that affect observations and environment dynamics: (i) rotation of observations by 90 degrees (Rotated), and (ii) color changes of background and items (Color-Swap).
The former preserves information prevalent in the environment, such as locations and color of items, while inducing a shift in the environment dynamics.
The latter entirely disregards information by altering item and background color.
We train our models for a total of $2\text{e}6$ interaction steps and apply an abrupt shift after every $n=5\text{e}5$ interaction steps, which would be sufficient for the algorithm to converge in the stationary setup.
Additionally, we test our algorithms in a custom version of the CartPole environment, in which we alter the environment dynamics.
For more details about the CartPole experiments, see Appendix \ref{appendixsub:cartpole}.

Unless stated otherwise, all experiments are conducted across 5 seeds, and the interquartile range (IQR) of the average reward over 1000 steps is shown.
Moreover, we perform statistical significance tests via a Wilcoxon rank-sum test \citep{wilcoxon_individual_1945} at a significance level of $\alpha_{wilcoxon} = 0.05$.
In this regard, we gather the average rewards after the first shift occurred and subsample every 10th value to perform pairwise comparisons between algorithms.

\subsection{Comparing PPO Against DQN Under Non-Stationarity}
\label{subsec:ppo_vs_dqn_nonstationarity}

\begin{figure}[hbt!]
    \centering
    \includegraphics[width=1.0\linewidth]{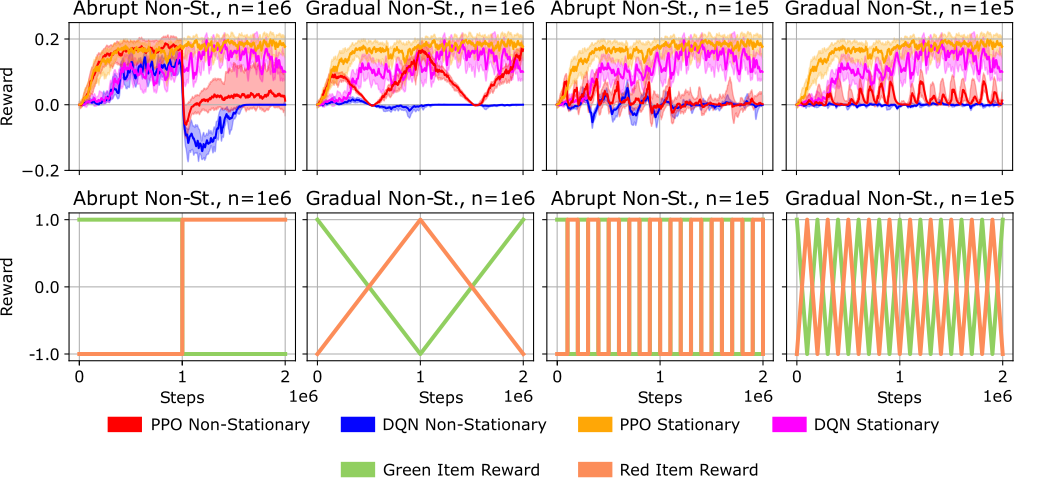}
    \caption{Non-stationary reward functions (bottom) with gradual or abrupt inversions, after or over the intervals (with $n=\{1\text{e}5,1\text{e}6$\}), respectively. Average rewards every 1000 steps are shown using an IQR and exponential average smoothing.}
    
    \label{fig:reward-functions}
\end{figure}

We conduct experiments to investigate how both PPO and DQN cope with non-stationarity in the reward function of JellyBean-Base, and which non-stationarities are particularly difficult to cope with.
As a point of reference, we additionally show the performance of PPO and DQN trained on the stationary version of JellyBean-Base.
The results for the different non-stationarities in the reward function are provided in Figure~\ref{fig:reward-functions}.
For a large interval of change and a gradual non-stationarity (2nd column), we observe that PPO is capable of adapting to the changing reward function, exhibiting promising forward transfer.
When the change occurs abruptly under the same interval, however (1st column), the performance of PPO deteriorates.
The most difficult setting is represented by an abrupt change and a small interval (3rd column), while results are slightly more stable for gradual changes and small intervals for PPO (4th column).
Nevertheless, PPO shows promising behavior in adapting to the different types of non-stationary reward functions.
In contrast to recent work which showed that the application of the softmax to policy gradient algorithms is error-prone in non-stationary environments \citep{hennes2020neural}, we find PPO to be quite stable.
A possible reason for that might be the constrained optimization strategy via trust regions, which enforces a small policy gap.
In contrast to PPO, DQN completely breaks regardless of the applied shift.
We additionally show the performance of a recurrent PPO agent in Figure \ref{fig:reward-shifts-all-lstm-ablation} on the reward shift experiments.
We find the recurrent agent to be much less sample efficient and prone to overfitting.

\subsection{Failure Modes of DQN}
\label{sec:dqn_failures}

As observed in previous experiments, DQN struggles to recover, especially when changes occur abruptly or very frequently.
By default, vanilla DQN uses $\epsilon$-greedy exploration with a decaying $\epsilon$, which results in a lower exploration rate in the later stages of training.
Furthermore, the replay buffer of DQN may still contain transitions collected under the previous stationarity regime after a domain shift has occurred.
Prior work has highlighted the importance of correctly re-using collected experiences in off-policy algorithms when non-stationarities are encountered~\citep{liotet2022lifelong,chandak2020optimizing,xie2021deep}.
In order to investigate the effect of the replay buffer and the exploration scheme, we conduct a thorough analysis on the components of DQN by:
\begin{enumerate}
\item reducing the replay buffer size with respect to the interval of change
\item injecting prior knowledge about when a change in the environment occurs into the $\epsilon$-greedy schedule (i.e. restarting exploration after a change occurred)
\item prioritizing samples with a high temporal difference (TD) error via prioritized experience replay (PER,~\citealp{schaul2016prioritized}) without bias correction
\item sampling actions from a distribution obtained via a softmax transformation of Q-values instead of the $\epsilon$-greedy schedule to add random exploration (DQN-stochastic)
\item measuring the effect of maximum entropy regularization via Soft Q-learning (SQL) \citep{haarnoja2017reinforcement}
\end{enumerate}

\begin{figure}[h]
\includegraphics[width=0.85\linewidth]{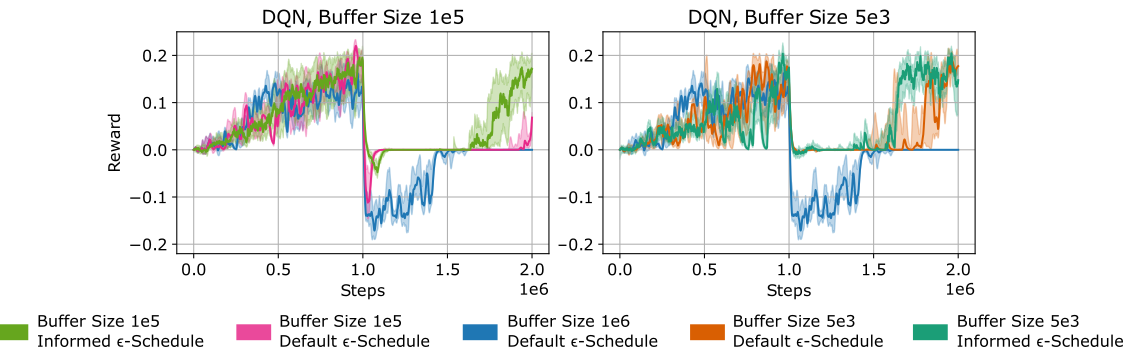}
\centering
\caption{Ablation on DQN buffer sizes ${1\text{e}5, 5\text{e}3}$, and exploration schedule under abrupt reward inversion after $n=1\text{e}6$. Average rewards every 1000 steps are shown using an IQR and exponential average smoothing.}
\label{fig:dqn_buffer_sizes}
\end{figure}

In Figure \ref{fig:non-stationary_dynamics-dqn-buffer-size-ablation} we show results for various replay buffer sizes on the Color Swap and Rotated environments. 
We find that flushing the replay buffer with new samples more frequently leads to improved results on Rotation, however it only leads to slightly better performance on Color Swap.
Also, reducing the buffer size by too much diminishes performance of DQN, indicating the existence of a sweet spot which depends on the interval of change.
Furthermore, if the interval of change is known a-priori, the exploration schedule may be adapted as well to restart exploration after a change in the environment occurred.
Figure \ref{fig:dqn_buffer_sizes} shows that with a reduced buffer size (left: $1\text{e}5$, and right: $5\text{e}3$) and adapted exploration schedule, DQN successfully recovers for abrupt shifts after $n=1\text{e}6$.
However, the information required for properly designing the exploration schedule and the replay buffer size is usually not available.
For implementation details, see Appendix \ref{appendixsub:dqn_impldetails}.

To alleviate the dependence on prior knowledge about the interval of change, we conduct additional experiments to improve upon vanilla DQN.
Naturally, after a domain shift occurs, newly sampled transitions will lead to a high TD error and can be prioritized via PER.
Initial experiments with PER indicate that the bias introduced via prioritized sampling can be beneficial to exploit within a domain.
However, across domains it rarely exhibits any gains (see DQN+PER, Figure \ref{fig:all-shifts-sql-ablation}).
Alternative exploration schemes such as DQN-stochastic do not provide any gains over DQN (see Figure \ref{fig:all-shifts-sql-ablation}).
Additionally, we compare DQN to SQL, which maximizes entropy while maximizing the return, and thus, is constantly encouraged to explore.
Figure \ref{fig:all-shifts-sql-ablation} shows that while SQL tends to improve over DQN after a shift in observations has occurred, it does not reach the performance of PPO.
This is especially apparent in environments with a low interval of change or gradually applied shifts in the rewards.
Also, these results highlight that even with the default buffer size, a more sophisticated exploration scheme can improve the performance over vanilla DQN without requiring external knowledge.
 
Our empirical evidence suggests that tuning the exploration strategy is a decisive factor for robustness in non-stationary environments when using DQN.
While injecting prior knowledge about the interval of change improves performance, such information is usually not available a-priori.
Even though PPO does not have access to this additional information, it outperforms DQN in environments with very frequent or gradual shifts.

\subsection{Breaking PPO with Non-Stationarity}
\label{subsec:breaking_ppo}

Although PPO is surprisingly resilient to changes in the reward function, we show that severe distribution shifts in the environment dynamics lead to failure cases.
Prior experiments have illustrated the capability of PPO to handle gradual shifts in a large interval of change.
Thus, in order to enforce failure cases of PPO, we focus exclusively on abrupt changes.

\begin{figure}[hbt!]
    \centering
    \includegraphics[width=0.75\linewidth]{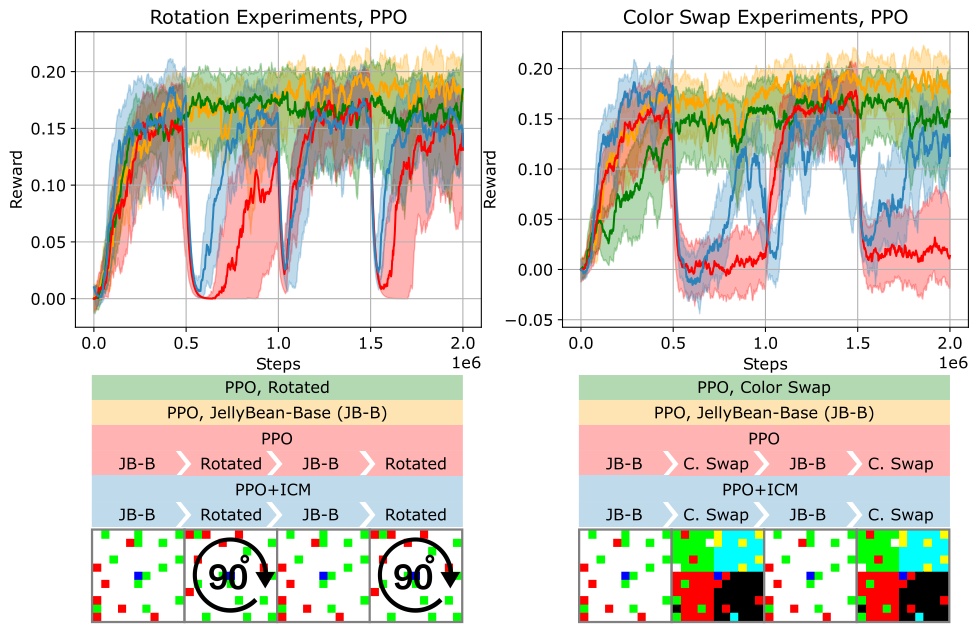}
    \caption{Experiments on non-stationary dynamics with PPO including PPO+ICM. Results are shown on the top row, while we visualize below when non-stationarity occurs due to Rotation or Color-Swap. Average rewards are shown every 1000 steps using an IQR and exponential average smoothing.}
    \label{fig:non-stationary_dynamics}
\end{figure}

Results for both, Rotated and Color-Swap non-stationarities, are shown in Figure \ref{fig:non-stationary_dynamics}.
As a baseline, we provide a comparison to a PPO agent trained in a stationary manner on JellyBean-Base and Color Swap/Rotation.
We find that in the isolated stationary cases,
both scenarios are easily solvable by PPO.
However, if they are combined to create a non-stationary scenario, PPO is experiencing difficulties. 
A change in the rotation of the observations does not cause too much harm, and PPO shows signs of forward transfer.
This might be due to the fact that non-rotated observations still contain useful information, such as locations of collectibles, that are transferred to the rotated case.
However, severely altering the background color and the color of the items disregards most information that would facilitate forward transfer, resulting in complete failure as soon as the stationarity changes.
Remarkably, as soon as the observations transition from the Color-Swap to the initial setting, PPO instantly recovers.
This indicates that no catastrophic forgetting occurs in our experimental setup.

\subsection{Fixing PPO with Reactive Exploration}
\label{subsec:fixing_ppo}

\begin{figure}[hbt!]
    \centering
    \includegraphics[width=0.75\linewidth]{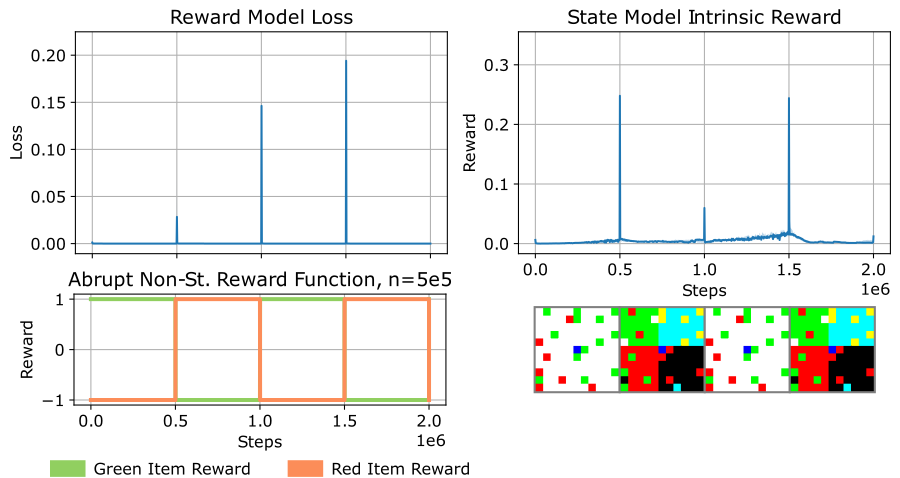}
    \caption{Results for changepoint detection via Reactive Exploration. On the left, the loss of the reward model (top) is depicted when trained on prediction of a non-stationary reward function (bottom).
    On the right, the loss of the novelty-based observation model (top) is shown for predicting the next observation (bottom). Reactive Exploration successfully detects various points of change. Reward (right) and loss (left) are shown every 1000 steps using an IQR and exponential average smoothing.}
    \label{fig:measuring-non-stationarity}
\end{figure}

Reactive Exploration helps detect environment changes and to cope with them.
Our previous experiments have shown that severe domain shifts considerably affect the performance of agents trained with PPO.
As discussed in Section~\ref{solution-proposal}, we incorporate Reactive Exploration in the form of intrinsic curiosity, and investigate its capabilities in detecting dynamically occurring changepoints.
Figure~\ref{fig:measuring-non-stationarity} shows the loss of the different components of the curiosity module.
We find that both, the reward model and the forward dynamics model, accurately detect changes in the reward function and in the observations, respectively.
The predictions of the forward dynamics model (right) appear to be slightly more noisy, which is expected since the task of predicting the next observation is inherently more difficult than predicting a single scalar value.
More details on the predicted reward of items are given in Section~\ref{appendixsub:reward-model-predictions}.
 
We repeat the Color-Swap experiment for which PPO failed, and incorporate the intrinsic rewards from the curiosity module.
Specifically, we set $\alpha = 0.85$ and $\beta = 0$ since no changes in the reward occur in this particular setup.
Equipped with Reactive Exploration, the agent manages to re-use previously acquired knowledge and quickly adapts to the new condition (see Figure~\ref{fig:non-stationary_dynamics} PPO+ICM).
PPO+ICM significantly outperforms PPO on the Rotation ($p=4.44\text{e}-9$) and Color-Swap environments ($p=3.44\text{e}-12$).
Experiments for additional hyperparameter settings can be found in Figure~\ref{fig:curiosity-results} and a discussion on our hyperparameter choices can be found in Appendix~\ref{appendixsubsub:hparams-reactive-exploration}. 
In general, too low values of $\alpha$ results in too little or no exploration at all, since the extrinsic rewards outweigh their intrinsic counterpart.
Vice-versa, too high values of $\alpha$ lead to exploration only.
Hence, to facilitate recovery, intrinsic and extrinsic rewards should be balanced to be roughly in the same range.
In Appendix \ref{appendixsub:cartpole} and Appendix \ref{appendixsub:reward-model-predictions}, we provide additional evidence on the usefulness of Reactive Exploration on a different environment under shifting environment dynamics, and on the reward-shift experiments presented in Section \ref{subsec:ppo_vs_dqn_nonstationarity}, respectively.

\section{Conclusion and Future Directions}
\label{sec:conclusion}

In \lifelong~RL an agent learns via interaction with a constantly evolving environment.
Such a scenario requires the agent to detect and cope with various domain shifts that arise over time, which most existing RL algorithms are not designed for.
In this paper, we propose to leverage Reactive Exploration to track and react to continual domain shifts in \lifelong~RL.
We conduct an empirical study comparing two popular RL algorithms, namely PPO and Q-learning, under a \lifelong~learning scenario when they face various types of non-stationarities.
We find that PPO is quite durable in the face of non-stationarity, while DQN drastically fails in the most simple cases, when only the reward function changes.
We conduct additional experiments to shed light on the failures of DQN, revealing that a more informative exploration mechanism can improve DQN.
For drastic changes in the observations, PPO eventually fails to recover after a shift is encountered.
As a remedy for such drastic shifts, we equip the PPO agent with Reactive Exploration in the form of an Intrinsic Curiosity Module, that helps to detect domain shifts, and to cope with them.
Equipped with Reactive Exploration, the agent successfully solves tasks under heavy domain shifts where performance has plummeted before, and shows promising signs of forward transfer. 
Overall, our analysis yields the insight that exploration mechanisms are vital for detecting and consequently dealing with continually occurring domain shifts.

Naturally, there is a multitude of ideas to extend our work.
Firstly, a more in-depth analysis of the effect of various non-stationarities on different algorithms potentially yields more insights on shortcomings and how to alleviate them.
Since our work draws a connection to the model-based RL literature, an interesting direction would be investigating how model-based algorithms perform in our framework.
Finally, building on our findings that stacking of various non-stationarities improves generalization (see Section~\ref{appendixsub:non-stationary-dynamics}) we would like to thoroughly investigate this effect.
\newpage
\section*{Acknowledgements}
The ELLIS Unit Linz, the LIT AI Lab, the Institute for Machine Learning, are supported by the Federal State Upper Austria. IARAI is supported by Here Technologies. We thank the projects AI-MOTION (LIT-2018-6-YOU-212), AI-SNN (LIT-2018-6-YOU-214), DeepFlood (LIT-2019-8-YOU-213), Medical Cognitive Computing Center (MC3), INCONTROL-RL (FFG-881064), PRIMAL (FFG-873979), S3AI (FFG-872172), DL for GranularFlow (FFG-871302), AIRI FG 9-N (FWF-36284, FWF-36235), ELISE (H2020-ICT-2019-3 ID: 951847). We thank Audi.JKU Deep Learning Center, TGW LOGISTICS GROUP GMBH, Silicon Austria Labs (SAL), FILL Gesellschaft mbH, Anyline GmbH, Google, ZF Friedrichshafen AG, Robert Bosch GmbH, UCB Biopharma SRL, Merck Healthcare KGaA, Verbund AG, Software Competence Center Hagenberg GmbH, T\"{U}V Austria, Frauscher Sensonic, AI~Austria~Reinforcement~Learning~Community and the NVIDIA Corporation.

\bibliography{refs}
\bibliographystyle{collas2022_conference}
\newpage
\appendix
\section*{\centering\huge\textbf{Supplementary Material}}
\vspace{2em}
\vskip -.8in
\addcontentsline{toc}{section}{Appendix}
\part{}
\parttoc
\section{Extended Results}
\label{appendix}

\subsection{DQN Failures - Experiment Details}
\label{appendixsub:dqn_impldetails}
In this section, we elaborate on the limiting factors of DQN in more detail. In particular, we conduct additional experiments with $\epsilon$-exploration schedules, reduced buffer sizes, Prioritized Experience Replay, Soft Q-Learning, and DQN-Stochastic.

\textbf{Vanilla DQN with reduced buffer size and $\epsilon$-exploration schedule.} We repeat the experiment with abrupt changes after a long interval of change of $n=1\text{e}6$ steps from Section~\ref{subsec:ppo_vs_dqn_nonstationarity}. In this setting, we add an $\epsilon$-exploration schedule in such a way that a second exploration surge is repeated exactly when the reward function changes in order to restart exploration.
A similar schedule is imposed by Reactive Exploration, as highlighted in Figure \ref{fig:measuring-non-stationarity}.

\begin{figure}[h]
  \centering
  \begin{subfigure}[b]{0.4\textwidth}
    \centering
    \includegraphics[width=\textwidth]{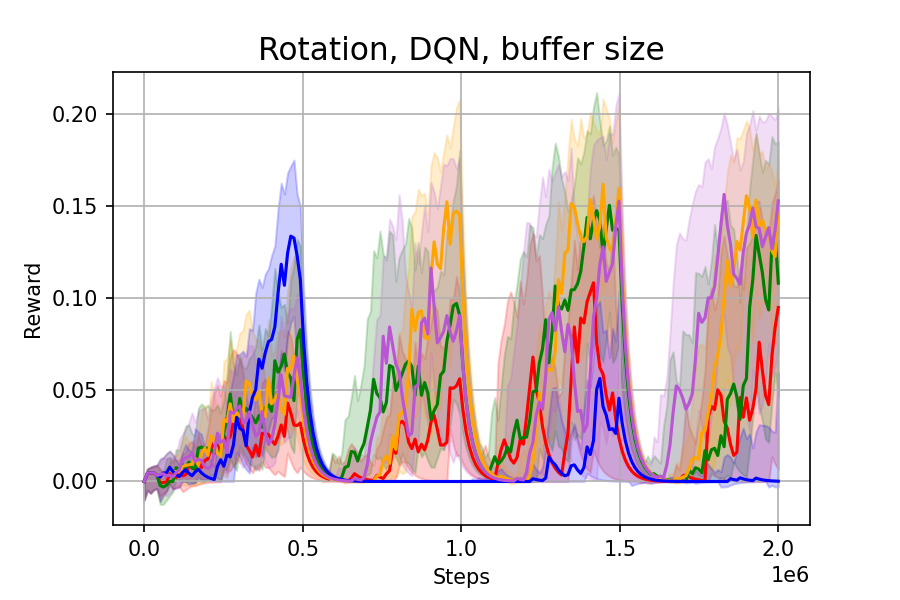}
  \end{subfigure}
  ~
  \begin{subfigure}[b]{0.4\textwidth}
    \centering
    \includegraphics[width=\textwidth]{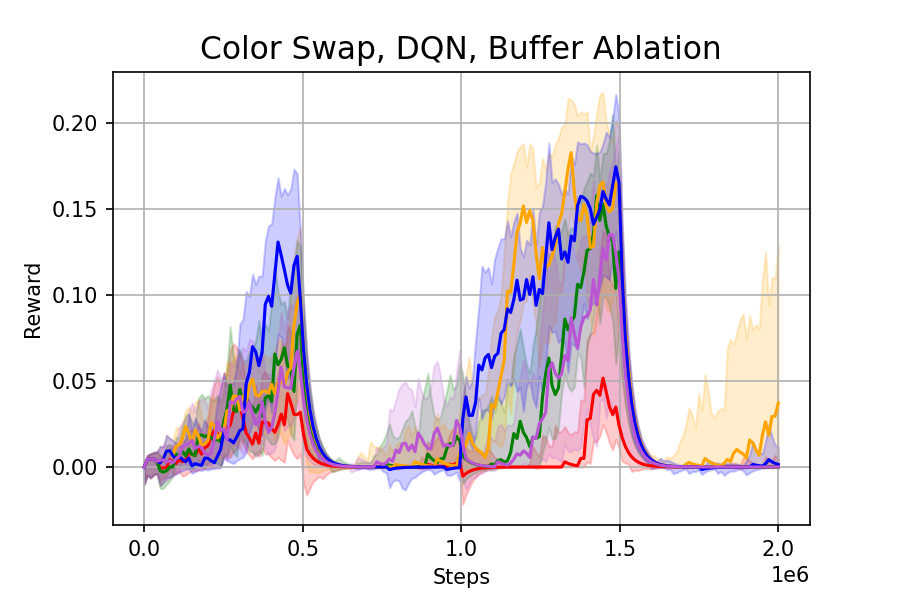}
  \end{subfigure}
    ~
  \begin{subfigure}[b]{0.8\textwidth}
    \centering
    \includegraphics[width=\textwidth]{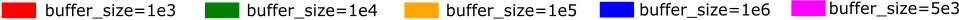}
  \end{subfigure}
  
  \caption{DQN buffer size ablation. Experiments for non-stationary dynamics. We investigate the effect using different buffer sizes with DQN. The default buffer size is 1e6.}
  \label{fig:non-stationary_dynamics-dqn-buffer-size-ablation}
\end{figure}

The default $\epsilon$-schedule is initialized at $\epsilon=1.0$, which resembles purely random behavior in the beginning, and decaying to $\epsilon=0.05$ over the first $10\%$ of interaction steps.
For the informative schedule, we repeat this $\epsilon$ schedule every time a domain shift occurs.
We call this schedule ''informed`` since it assumes prior knowledge about the occurring shifts.
Additionally, we limit the buffer size to contain either $1e5$, or $5e3$ sampled transitions.
We create four experiments 
for all combinations of default versus informed $\epsilon$-schedule and buffer size $1e5$ versus $5e3$, and contrast them to the default $1e6$ with the default $\epsilon$-schedule.
The results of these experiments can be seen in Figure~\ref{fig:dqn_buffer_sizes}.
In addition, we provide an ablation over the buffer sizes $\{1\text{e}3, 5\text{e}3, 1\text{e}4, 1\text{e}5 \}$ for the observation shift environments in Figure~\ref{fig:non-stationary_dynamics-dqn-buffer-size-ablation}.
Reducing the buffer size results in quick recovery in the Rotation environment as compared to the default buffer size of $1\text{e}6$.
In the Color-Swap environment, a smaller buffer results in marginal gains, but a promising trend in terms of recovery.

\begin{figure}[h]
  \centering
  \begin{subfigure}[b]{0.4\textwidth}
    \centering
    \includegraphics[width=\textwidth]{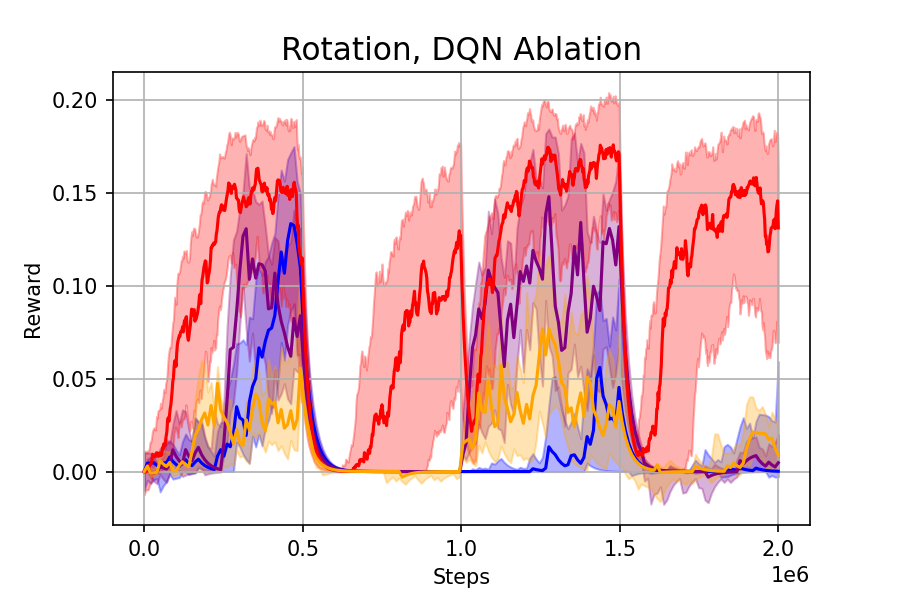}
  \end{subfigure}
  ~
  \begin{subfigure}[b]{0.4\textwidth}
    \centering
    \includegraphics[width=\textwidth]{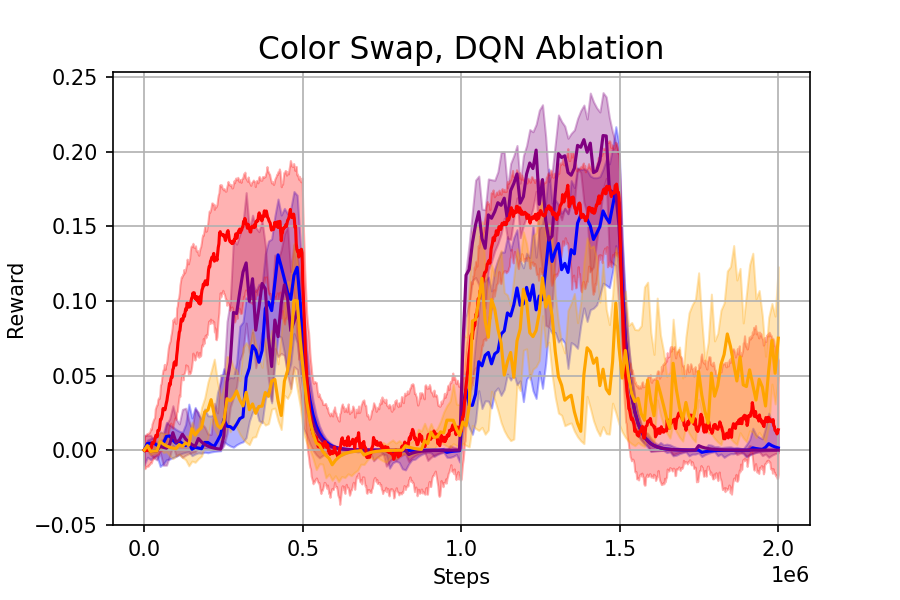}
  \end{subfigure}
    ~
  \begin{subfigure}[b]{0.4\textwidth}
    \centering
    \includegraphics[width=\textwidth]{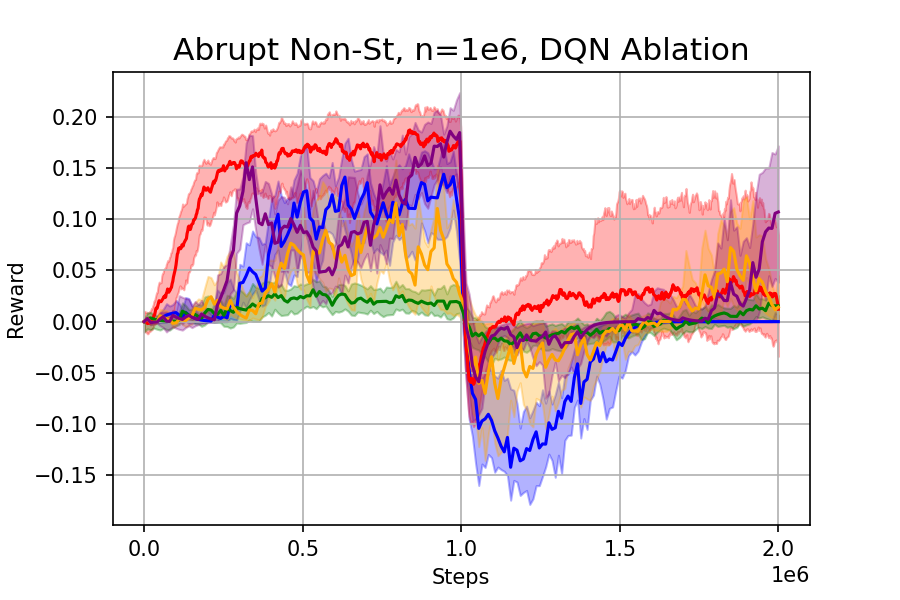}
  \end{subfigure}
  ~
  \begin{subfigure}[b]{0.4\textwidth}
    \centering
    \includegraphics[width=\textwidth]{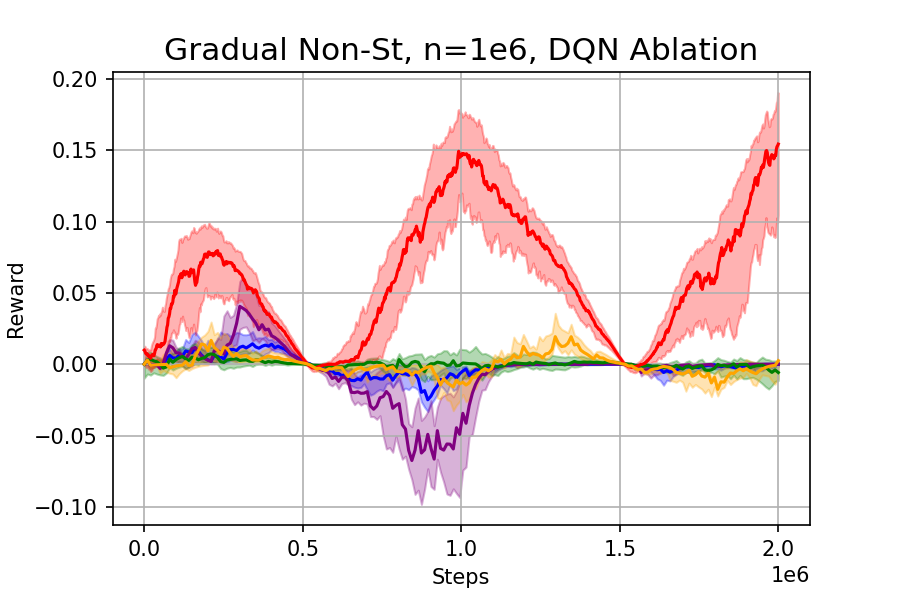}
  \end{subfigure}
    ~
  \begin{subfigure}[b]{0.4\textwidth}
    \centering
    \includegraphics[width=\textwidth]{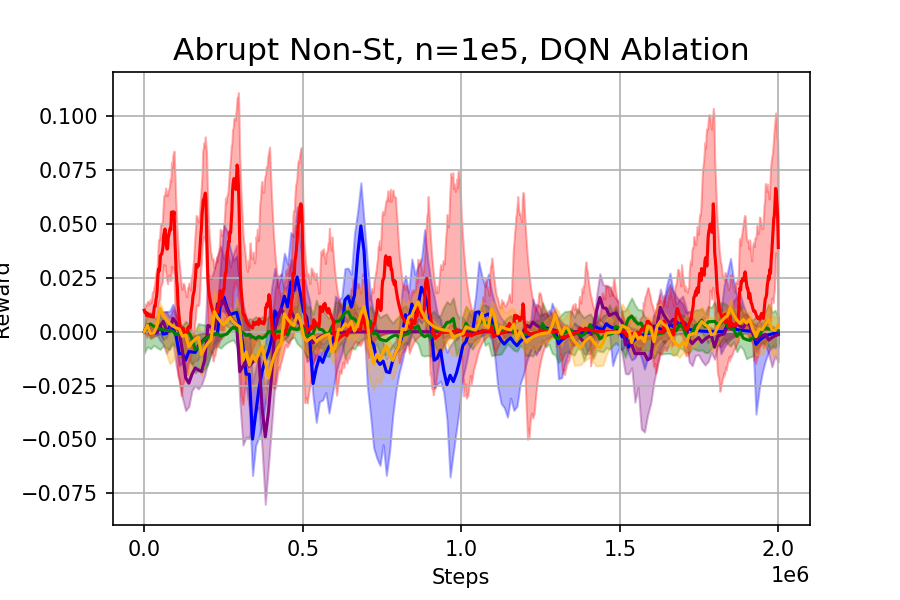}
  \end{subfigure}
      ~
  \begin{subfigure}[b]{0.4\textwidth}
    \centering
    \includegraphics[width=\textwidth]{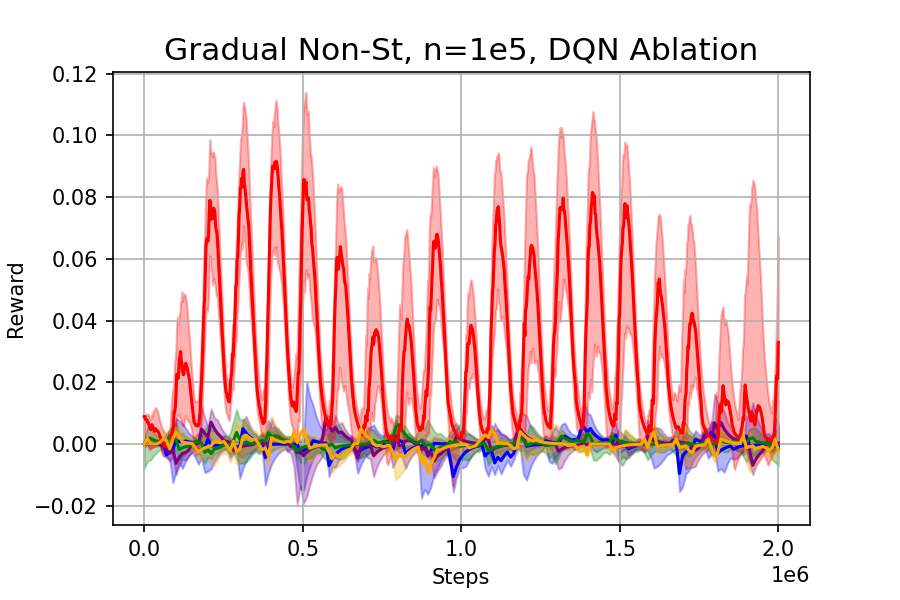}
  \end{subfigure}
  ~
  \begin{subfigure}[b]{0.6\textwidth}
    \centering
    \includegraphics[width=\textwidth]{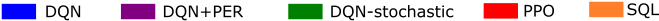}
  \end{subfigure}
  \caption{Experiments for non-stationary dynamics (top row) and non-stationary reward functions (middle, bottom row). We investigate the effect of DQN+PER and SQL and compare them to the PPO and DQN baselines.}
  
  \label{fig:all-shifts-sql-ablation}
\end{figure}

\textbf{DQN + PER.} We use the default value for $\alpha_{PER}=0.6$. 
Our initial suspicion is that the bias introduced via prioritization of samples with a high TD-error might have a positive impact in the non-stationary setup.
This is due to the fact that transitions after a shift will exhibit a higher TD-error than others in the buffer and will be prioritized.
Therefore, we set $\beta_{PER} = 0$. 
We show in Figure \ref{fig:all-shifts-sql-ablation} that while PER improves performance for one type of stationarity, agents only recover rarely and there is no significant improvement upon DQN in terms of recovery.

\textbf{Soft Q-Learning} In Figure \ref{fig:all-shifts-sql-ablation} we additionally show the learning curve for SQL on all environments. 
SQL maximizes entropy while maximizing the return and, thus, is constantly encouraged to explore the environment. 
As described in Section \ref{sec:dqn_failures}, 
While SQL tends to improve over DQN after a shift in observations has occurred, it does not reach the performance of PPO.

\textbf{Stochastic DQN.} Figure \ref{fig:all-shifts-sql-ablation}, presents the performance of DQN-stochastic.
DQN-stochastic is a form of randomized exploration which is enforced via a probability distribution over predicted Q-values. 
We observe that this results in inferior performance due to small differences between Q-values, which results in an almost uniform action distribution, and thus almost fully random exploration and little to no exploitation.

\subsection{Repeating Experiments in CartPole}
\label{appendixsub:cartpole}
We repeat several of our experiments using a custom \lifelong~version of the popular CartPole environment, in which we can easily induce non-stationarity in environment dynamics. 
First, we elaborate on the environment, followed by experiments with PPO and DQN. 
Further, we design difficult non-stationarity to break PPO, and finally, show promising behavior in terms of recovery after adding Reactive Exploration.

In the Cartpole environment, the agent must balance a pole which is place on a cart that it can freely move.
Usually, the environment is reset after the pole leans towards one side with a certain angle.
We reset the pole when it drops, but allow the cart to move infinitely in a single episode.
In this regard, we set the cart's velocity and acceleration to $0$ when it hits the edges of the box, which turns the problem to an infinite horizon problem. 
This setup resembles the most similar to the episodic case, since restricting the cart movement inside its box prevents a velocity buildup that would not be possible in the episodic setting.

In addition, we apply a variable reward function \citep{regmi2020infinite}, in which a perfectly balanced pole yields $+0.5$ reward, and the perfectly centered cart yields $+0.5$ reward, totaling $+1.0$. A sum of $0$ reward is yielded exactly at the point at which both the cart position $x$ and the pole angle $\theta$ are at their original thresholds. If the pole drops further, the reward drops drastically below $0$. 
Formally, the reward function is defined as: 
\begin{align*}
    r=1 - 11.52 \cdot x^2 - \frac{\theta^2}{288} 
\end{align*}

Creating settings with modified dynamics can be achieved by changing the properties of the environment, which are based on physics calculations. 
The default settings, which we use as a baseline for comparison against our modifications, are defined by a gravity of $9.8$, a cart mass of $1.0$, a pole mass of $0.1$, a pole length of $0.5$, and a force magnitude of $10.0$.
The sign of the force magnitude controls the direction of the cart movement.

\begin{figure}[hbt!]
    \centering
    \includegraphics[width=0.75\linewidth]{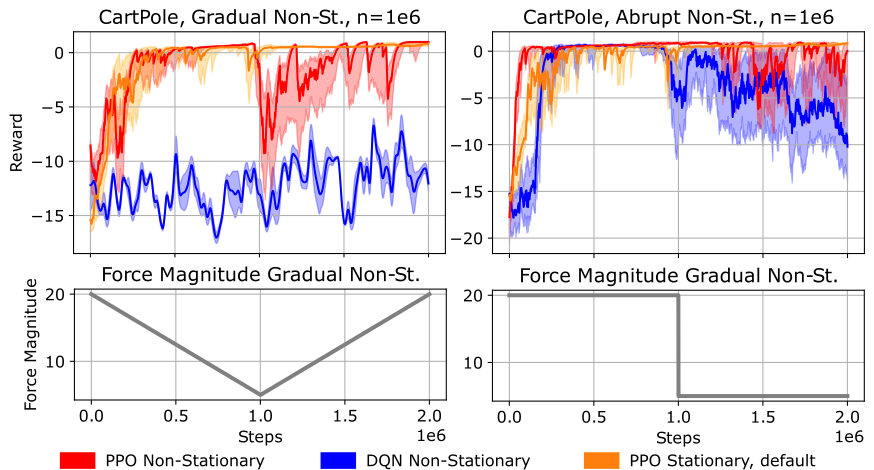}
    \caption{Experiments for non-stationary dynamics in CartPole, comparing DQN and PPO agents. The top shows the results, and below we visualize how the non-stationary dynamics occur. Average rewards are logged every 1000 steps and shown using an IQR and exponential average smoothing.}
    \label{fig:cartpole_ppo_vs_dqn}
\end{figure}

We create abrupt and gradual non-stationarity in the environment dynamics.
To this end, we change the force magnitude from $20$ to $5$ either gradually over $1\text{e}6$ steps and back again over the next $1e6$ steps, or abruptly every $1\text{e}6$ steps.
Again, we add the performance of  PPO and DQN agents trained on the stationary environment under default parameters as a baseline (see Figure \ref{fig:cartpole_ppo_vs_dqn}).
In the results for the gradual shift (left), we can observe how DQN is not capable of solving the task at all.
This is likely due to the fact that adapting to a changing force magnitude in order to keep the balance is too difficult with outdated knowledge in the replay buffer.
PPO, on the other hand, manages to learn how to balance the pole quite well. 
However, we can observe that once the force starts to increase instead of decrease, PPO agents become more unstable.
In the abrupt case (right), both algorithms initially learn to balance as expected. 
Once the environment dynamics change, PPO agents adapt well with minor instabilities, while DQN again struggles more severely.

\begin{figure}[h]
    \centering
    \includegraphics[width=0.9\linewidth]{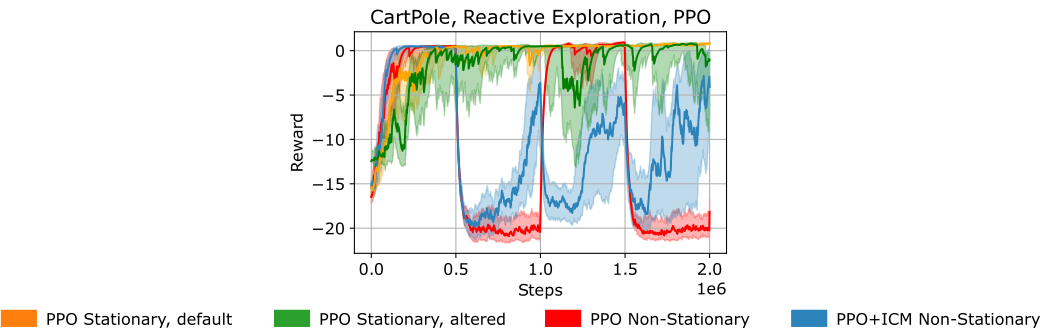}
    \caption{Results for experiments on difficult non-stationarity in CartPole to break PPO and Reactive Exploration to alleviate the difficulties. Average rewards every 1000 steps are shown using an IQR and exponential average smoothing. For the default settings, we use a cart mass of 1.0, a pole length
of 0.5, and a force magnitude of 10.0. When switching to the altered setting, these turn into a pole length of $1.0$, a cart mass of $2.0$, and a force magnitude of $-10.0$ (inverting the direction of movement).}
    \label{fig:cartpole_breaking_ppo_curiosity}
\end{figure}

Next, we explore domain shifts that reliably render PPO unable to recover. 
As we experienced in prior experiments, simple modifications such as increasing or decreasing the force magnitude only introduces minor instabilities.
We instead focus on changing the sign of the force magnitude, resulting in an inversion between left and right movement of the cart. 
Additionally, we set the environment parameters to a pole length of $1.0$, and a cart mass of $2.0$.
Further, we apply the same interval of change as used in our JBW experiments at $5\text{e}5$ steps until force magnitude is inverted. 
A stationary baseline is trained on the default as well as the altered dynamics and shown as a reference. 
The results show that both the default and the new dynamics are learnable in the stationary case, but when switching between them, PPO is unable to recover (see Figure \ref{fig:cartpole_breaking_ppo_curiosity}, red line).

Finally, we apply our Reactive Exploration, including the forward and inverse dynamics models, as well as the reward model.
For scaling of the intrinsic reward, we observe the magnitude of the model rewards in early experiments, scale down $r_t^r$ by a factor of $0.2$, and set the parameters $\alpha=0.15$, and $\beta=0.15$. 
Results (see Figure~\ref{fig:cartpole_breaking_ppo_curiosity}, blue line) show that agents equipped with Reactive Exploration successfully recover after encountering the shift in the environment dynamics.
For additional details on hyperparameter selection, see Appendix \ref{appendix:experiment-setup}.

\subsection{Reward Model Predictions}
\label{appendixsub:reward-model-predictions}
\begin{figure}[h]
    \centering
    \includegraphics[width=0.75\linewidth]{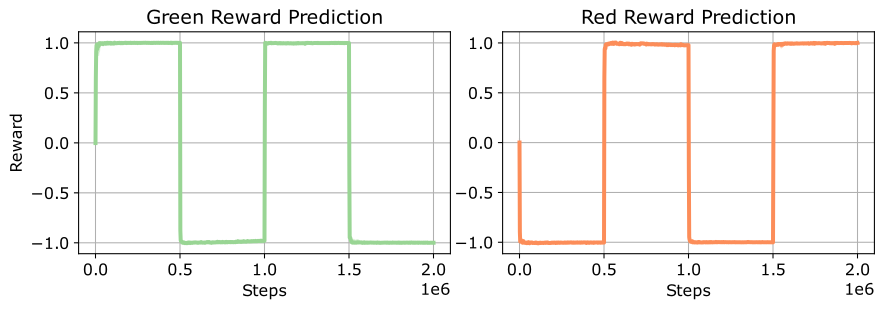}
    \caption{Predicted reward for picking up green and red reward using the reward model, which perfectly represent the true rewards yielded by green and red items as they are inverted every $n=5\text{e}5$ steps between $+1.0$ and $-1.0$. Average rewards every 1000 steps are shown using an IQR and exponential average smoothing.}
    \label{fig:reward_model_predictions}
\end{figure}

We visualize the predictions of the reward model for the abrupt inversion of the reward function every $n=5\text{e}5$ steps (Section \ref{subsec:fixing_ppo}).
Figure \ref{fig:reward_model_predictions} shows that the reward model almost perfectly predicts the true reward function, as the model immediately adapts to the changes.
Thus, a reward model can effectively detect and respond to non-stationarities in the reward function.

\clearpage
\subsection{Reactive Exploration for Reward Shift Experiments}
\begin{figure}[h]
  \centering
  \begin{subfigure}[b]{0.4\textwidth}
    \centering
    \includegraphics[width=\textwidth]{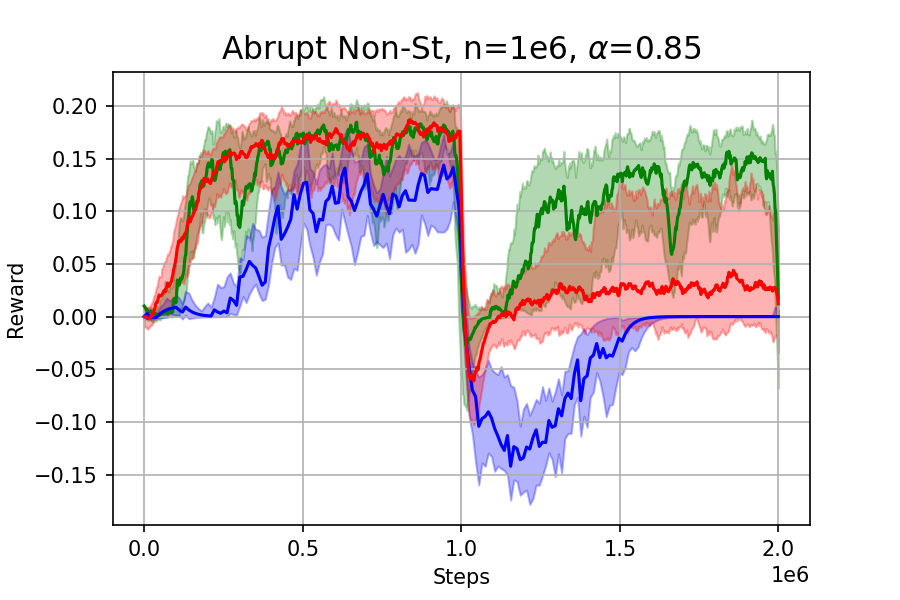}
  \end{subfigure}
  ~
  \begin{subfigure}[b]{0.4\textwidth}
    \centering
    \includegraphics[width=\textwidth]{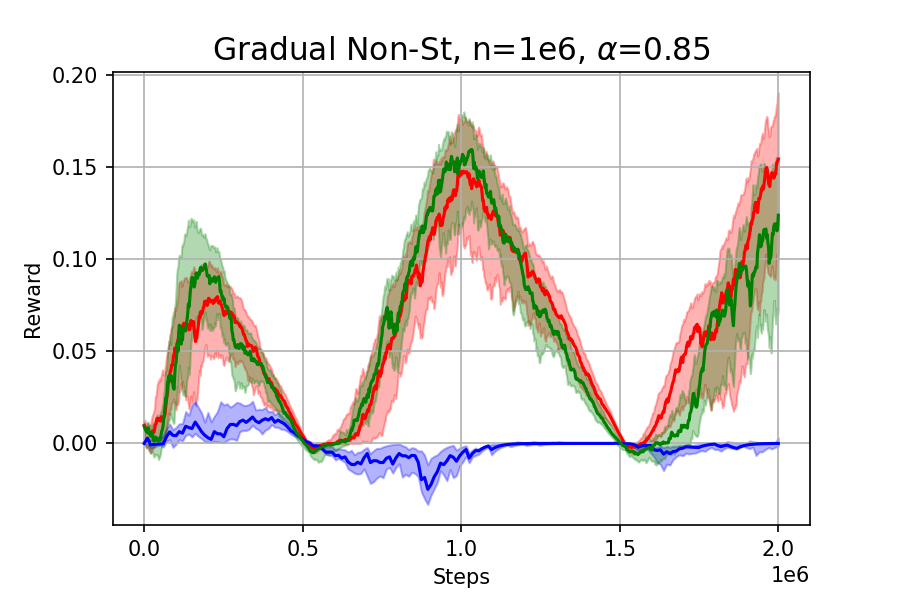}
  \end{subfigure}
    ~
  \begin{subfigure}[b]{0.4\textwidth}
    \centering
    \includegraphics[width=\textwidth]{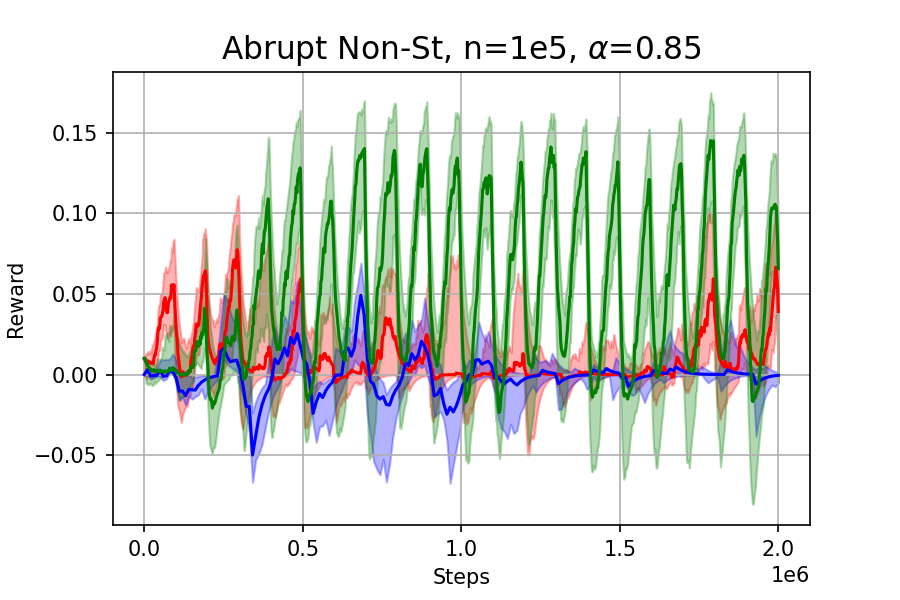}
  \end{subfigure}
      ~
  \begin{subfigure}[b]{0.4\textwidth}
    \centering
    \includegraphics[width=\textwidth]{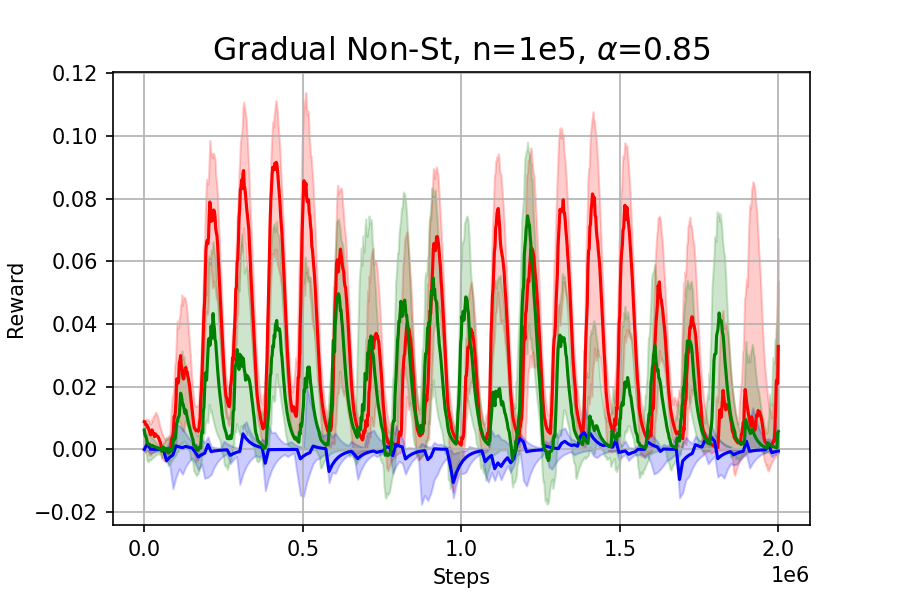}
  \end{subfigure}
  ~
  \begin{subfigure}[b]{0.28\textwidth}
    \centering
    \includegraphics[width=\textwidth]{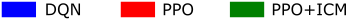}
  \end{subfigure}
 
  \caption{Experiments for non-stationary reward functions with abrupt (left) or gradual (right) reward shifts. We compare the baselines PPO and DQN to PPO with Reactive Exploration (ICM, $\alpha=0.85$).} 
  \label{fig:reward-shifts-all-bestalpha}
\end{figure}

In Figure \ref{fig:non-stationary_dynamics} (Section \ref{subsec:breaking_ppo}), we showed how Reactive Exploration mitigates the impact of observation shifts on the agent's performance. 
For completeness, we provide experiments with Reactive Exploration on environments with reward shifts (Figure \ref{fig:reward-functions}). 
In Figure \ref{fig:reward-shifts-all-bestalpha}, we show the learning curves on abrupt (left) and gradual (right) reward shifts over $n=1\text{e}6$ (top) and $n=1\text{e}5$ (bottom) steps for the agent equipped with Reactive Exploration (PPO+ICM, green) and compare to PPO (red) and DQN (blue) baselines. 
For all four tasks, we use an intrinsic reward weight $\alpha=0.85$ for PPO+ICM.

On experiments comprising abrupt shifts, PPO+ICM significantly outperforms PPO for an interval of change of $n=1\text{e}6$ (top left, $p=8.28e-9$). 
Even for frequently occurring reward shifts ($n=1\text{e}5$, bottom left), PPO+ICM significantly outperforms PPO ($p=7.87e-9$).
On gradual shifts, PPO without exploration already performs close to optimal.
Thus, there is not much gain in adding Reactive Exploration.

\subsection{Recurrent PPO Ablation}
\label{app:rec_ppo}
\begin{figure}[h]
  \centering
  \begin{subfigure}[b]{0.4\textwidth}
    \centering
    \includegraphics[width=\textwidth]{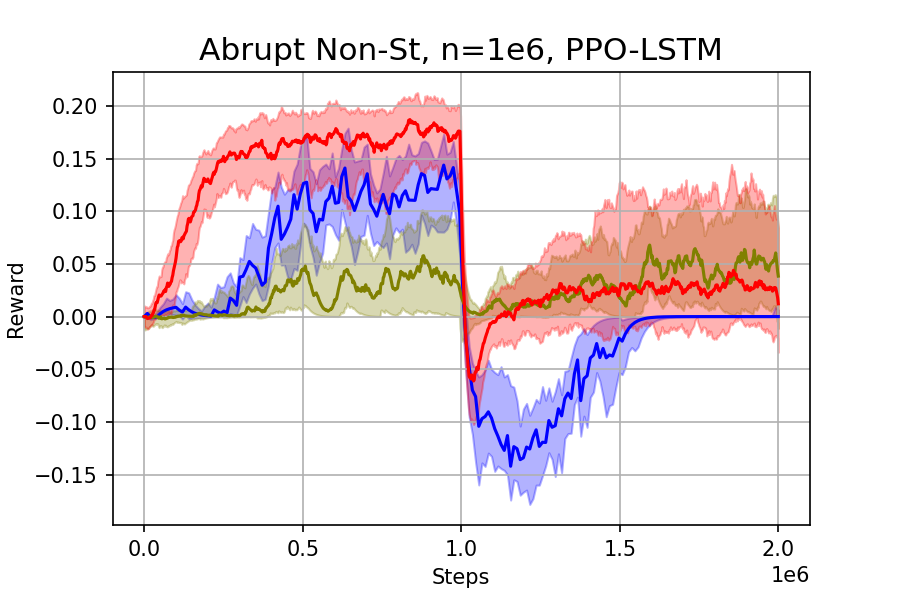}
  \end{subfigure}
  ~
  \begin{subfigure}[b]{0.4\textwidth}
    \centering
    \includegraphics[width=\textwidth]{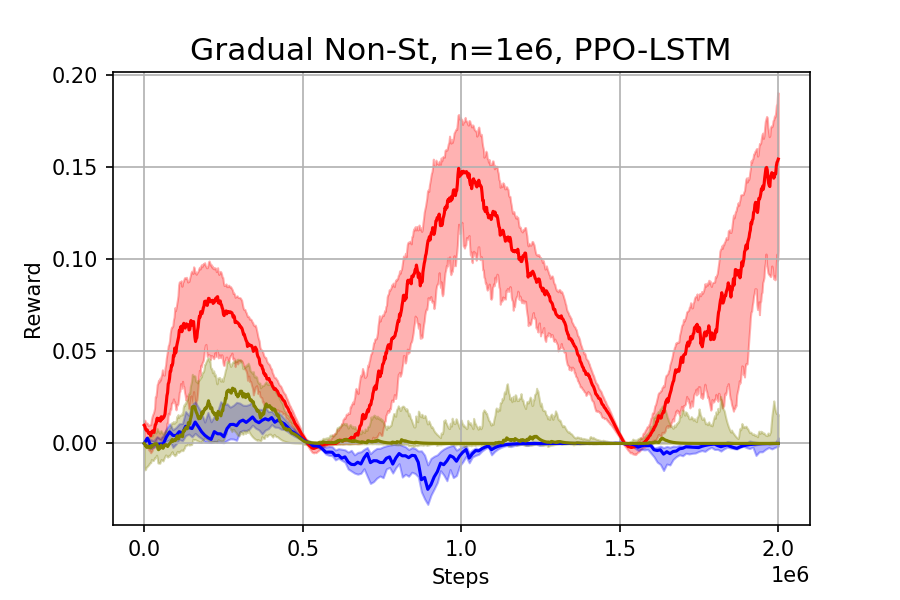}
  \end{subfigure}
    ~
  \begin{subfigure}[b]{0.4\textwidth}
    \centering
    \includegraphics[width=\textwidth]{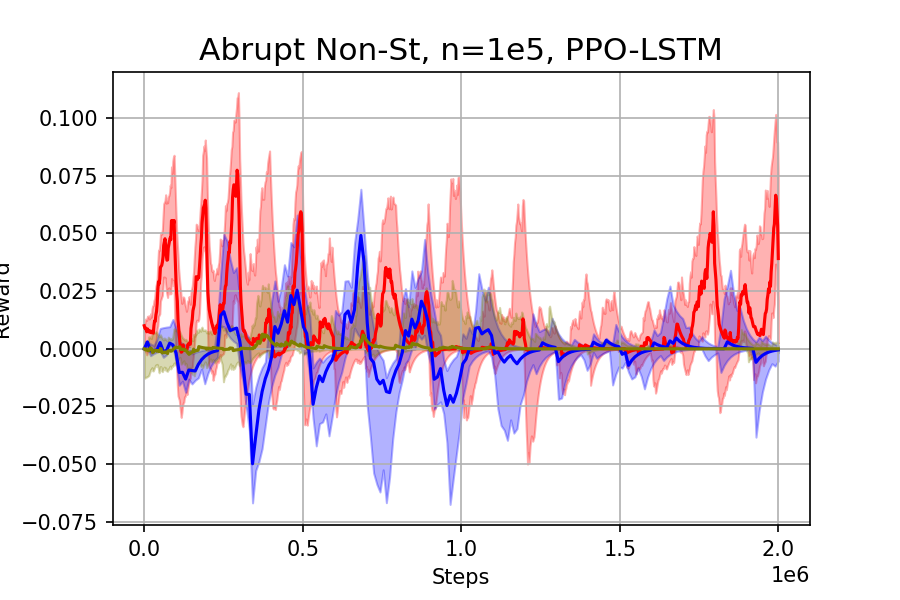}
  \end{subfigure}
      ~
  \begin{subfigure}[b]{0.4\textwidth}
    \centering
    \includegraphics[width=\textwidth]{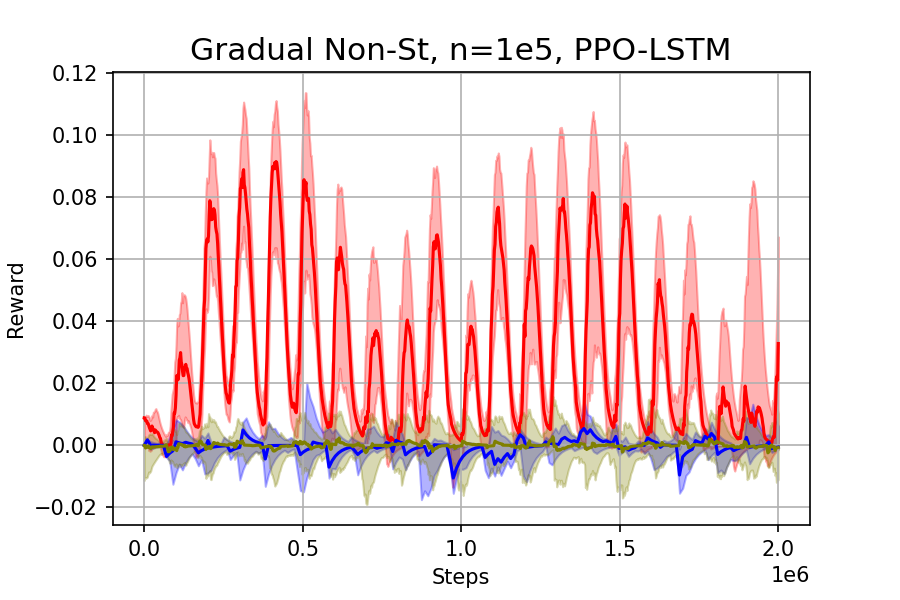}
  \end{subfigure}
  ~
  \begin{subfigure}[b]{0.35\textwidth}
    \centering
    \includegraphics[width=\textwidth]{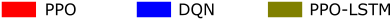}
  \end{subfigure}
   \caption{PPO-LSTM ablation. Experiments for non-stationary reward functions with abrupt (left) or gradual (right) reward shifts. We compare the baselines PPO and DQN to PPO-LSTM.}
  \label{fig:reward-shifts-all-lstm-ablation}
\end{figure}

To investigate the robustness of a recurrent architecture to the different domain shifts, we add an experiment where we use an LSTM \citep{hochreiter1997long} on top of the convolutional backbone (PPO-LSTM). 
Since the task the agent is required to solve is not dependent on memory, we sample sequence chunks of length 8 and reset hidden and cell state for computing the gradients. 
Figure \ref{fig:reward-shifts-all-lstm-ablation} shows the performance of PPO-LSTM compared to DQN and PPO.
We observe that a recurrent agent is much less sample efficient than a Markovian policy.
This is clearly visible for an abrupt shift and a large interval of change ($n=1\text{e}6$), where PPO-LSTM shows a promising trend in terms of recovery after being trained sufficiently long (top left).
However, for gradual shifts (right column) or frequent shifts ($n=1\text{e}5$, bottom left) PPO-LSTM performs poorly.
This is due to the fact that it frequently overfits to avoid collecting any item, since it is not capable of adapting fast enough to the new stationarity.

\subsection{Possible Instantiations for Reactive Exploration}
\label{appendixsub:other-exploration-methods}
As mentioned in Section \ref{solution-proposal} we instantiate our Reactive Exploration based on ICM \citep{pathak2017curiosity}.
However, any other exploration mechanism suited for the infinite horizon setup may be used.
In this regard, we investigate different instantiations of Reactive Exploration.
\begin{figure}[H]
  \centering
    \begin{subfigure}[b]{0.4\textwidth}
    \centering
    \includegraphics[width=\textwidth]{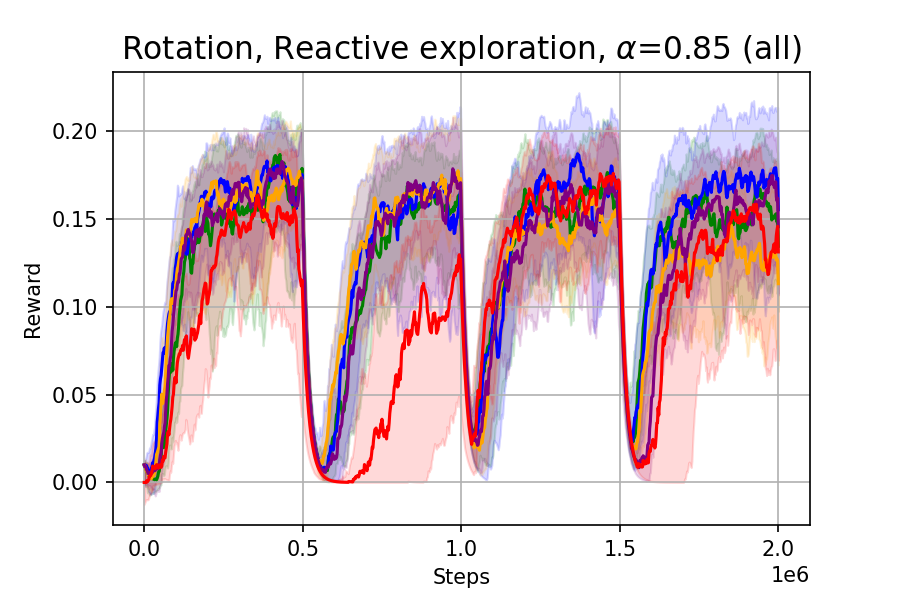}
  \end{subfigure}
  ~
  \begin{subfigure}[b]{0.4\textwidth}
    \centering
    \includegraphics[width=\textwidth]{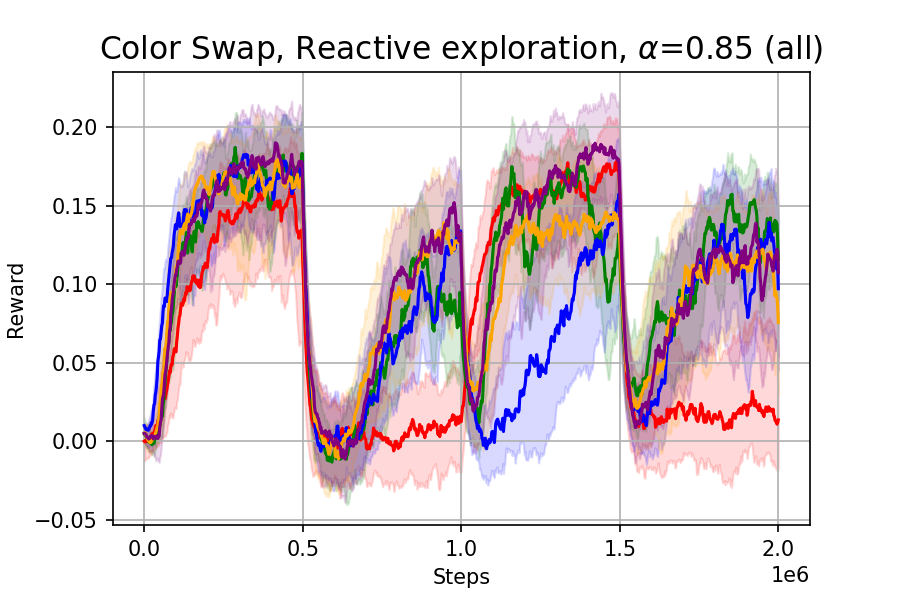}
  \end{subfigure}
  
  \begin{subfigure}[b]{0.4\textwidth}
    \centering
    \includegraphics[width=\textwidth]{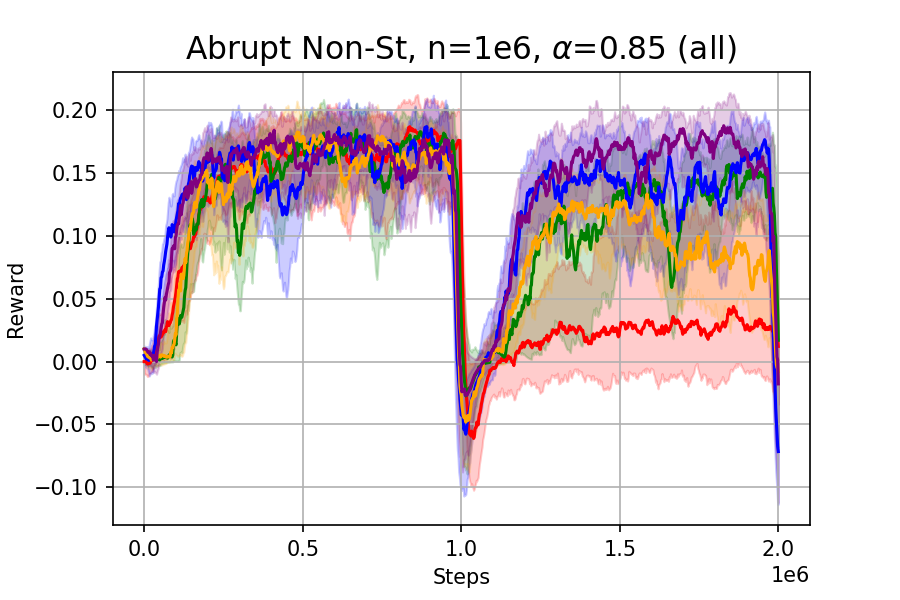}
  \end{subfigure}
  ~
  \begin{subfigure}[b]{0.4\textwidth}
    \centering
    \includegraphics[width=\textwidth]{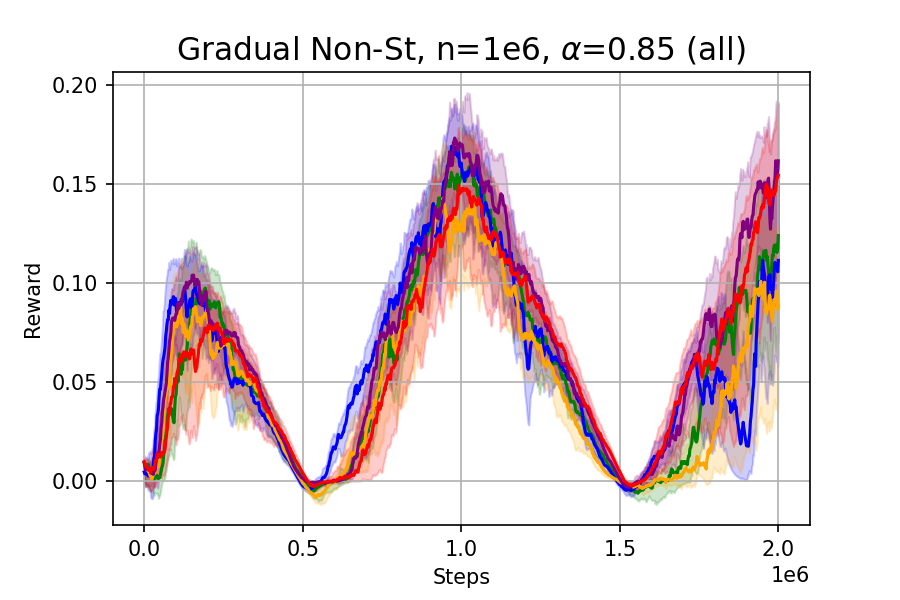}
  \end{subfigure}
    ~
  \begin{subfigure}[b]{0.5\textwidth}
    \centering
    \includegraphics[width=\textwidth]{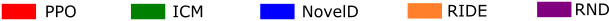}
  \end{subfigure}
  \caption{Experiments for non-stationary dynamics (top row) and non-stationary reward functions (bottom row). We investigate the effect of three additional exploration methods (RND, RIDE and NovelD) and compare them to the PPO baseline. We use $\alpha=0.85$ for all experiments.}
  \label{fig:state-and-reward-shifts-with-all-exp}
\end{figure}

Specifically, we consider three other prominent exploration methods: RND \citep{burda2019exploration}, RIDE \citep{raileanu2020ride}, and NovelD \citep{zhang2021noveld}. 
Similar to ICM, these three additional methods assign a curiosity-based reward bonus for exploration. 
RND maintains two networks that receive observations from the environment: (i) a fixed, randomly initialized target network and (ii) a predictor network, that aims to predict the outputs of the target network. 
The intrinsic reward is computed as the prediction error of the predictor network. 
Unlike ICM and RND, both RIDE and NovelD compute the intrinsic rewards as the novelty difference of two consecutive states. 
To quantify the novelty of states, RIDE and NovelD rely on ICM and RND, respectively. 
In contrast to ICM and RND, however, RIDE and NovelD were designed for episodic environments and maintain a lookup table of state visitation counts per episode. 
RIDE normalizes the exploration bonus by the root of the episodic state visitation count, and NovelD only assigns the intrinsic reward the first time a state was visited in the current episode. 
As such, in their original form, they cannot be applied to infinite horizon problems. 
Therefore, we make slight adjustments, and only maintain lookup tables per rollout of PPO (2048 steps), not per episode.

In Figure \ref{fig:state-and-reward-shifts-with-all-exp} we present results for different instances of Reactive Exploration. 
We conduct experiments on two environments with observation-shifts (Rotation and Color-Swap, top row) as well as two environments with reward shifts (abrupt/gradual shift after/over 1e6 steps, bottom row). 
In line with experiments presented in Section \ref{appendixsubsub:hparams-reactive-exploration}, all agents use the intrinsic reward weight $\alpha=0.85$. 
We show the learning curves over eight seeds and compare them to the PPO baseline. 
Overall, we observe that RND (purple), RIDE (yellow) and NovelD (blue) perform on-par with our instantiation of ICM. 
Notably, the only method that significantly outperforms the PPO baseline on gradual reward shift (bottom right) is RND ($p=1.02\text{e}-5$).
On all other experiments, all methods significantly outperform the PPO baseline.
These results suggest, that other state-of-the-art exploration mechanisms may be used for Reactive Exploration in the \lifelong\ RL setup. 

\subsection{Comparing Reactive Exploration and Entropy Maximization}
To shed more light on what types of exploration are most efficient in facilitating recovery of PPO, we conduct another set of experiments.
Particularly, we compare our curiosity-based exploration (PPO+ICM) against the standard maximum entropy regularization of PPO (PPO+ME) with $\epsilon=0.01$.
In addition, we show reference results for PPO, and for combining both Reactive Exploration via ICM and entropy regularization (PPO+ICM+ME).

\begin{figure}[h]
  \centering
  \begin{subfigure}[b]{0.4\textwidth}
    \centering
    \includegraphics[width=\textwidth]{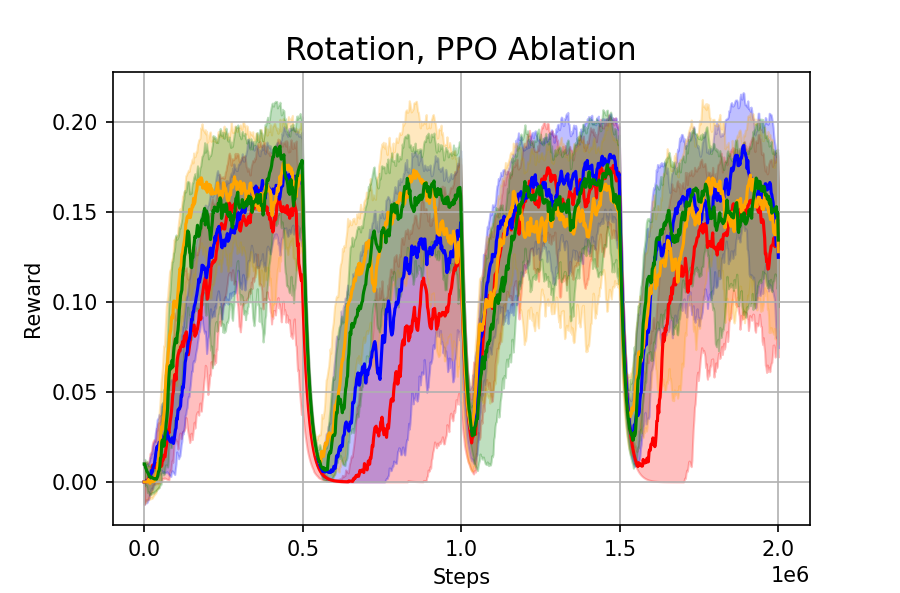}
  \end{subfigure}
  ~
  \begin{subfigure}[b]{0.4\textwidth}
    \centering
    \includegraphics[width=\textwidth]{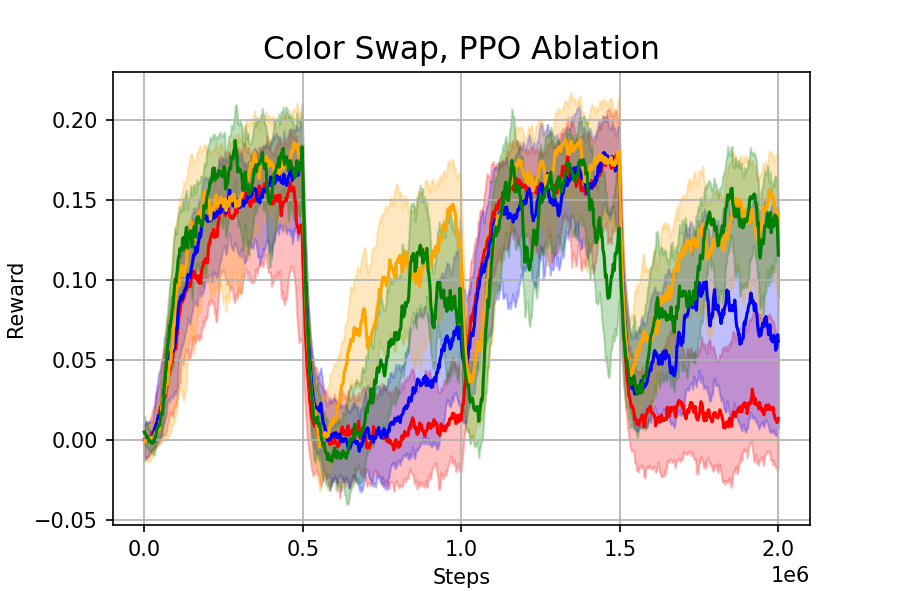}
  \end{subfigure}
    ~
  \begin{subfigure}[b]{0.4\textwidth}
    \centering
    \includegraphics[width=\textwidth]{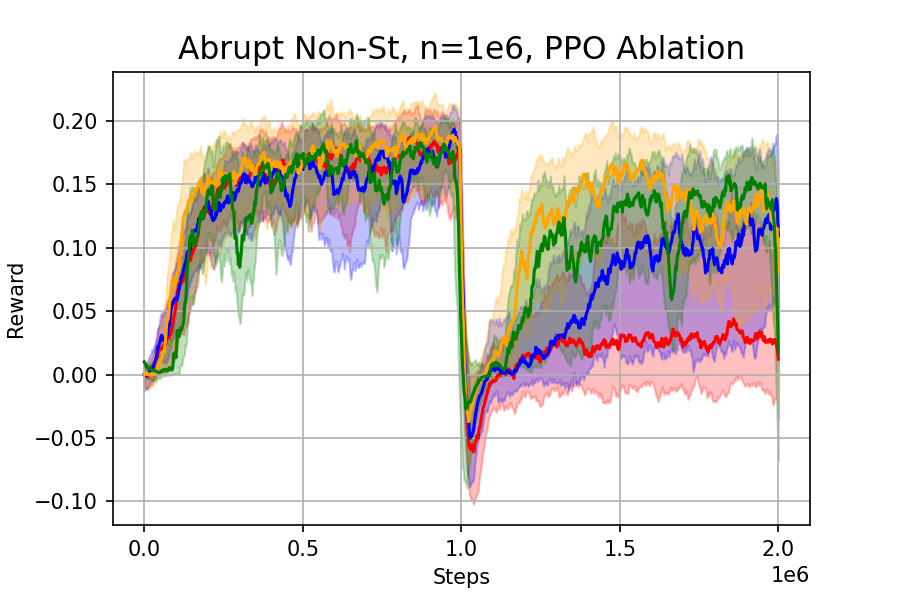}
  \end{subfigure}
  ~
  \begin{subfigure}[b]{0.4\textwidth}
    \centering
    \includegraphics[width=\textwidth]{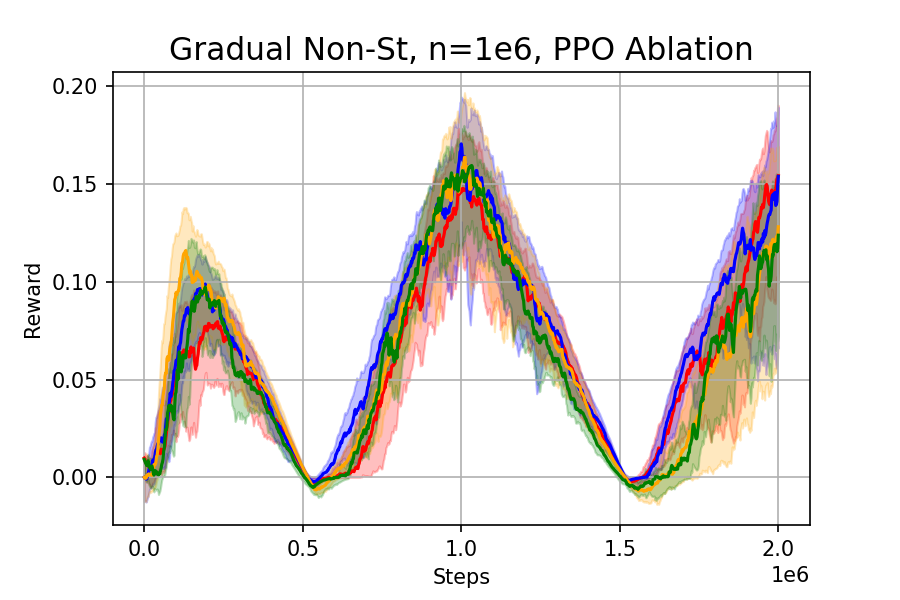}
  \end{subfigure}
  ~
  \begin{subfigure}[b]{0.53\textwidth}
    \centering
    \includegraphics[width=\textwidth]{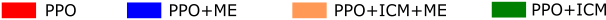}
  \end{subfigure}
  \caption{Experiments for non-stationary dynamics (top row) and non-stationary reward functions (bottom row). We compare the effect of PPO+ME and PPO+ICM+ME and to PPO+ICM and the PPO baseline.}
  
  \label{fig:all-shifts-maxent}
\end{figure}

Figure \ref{fig:all-shifts-maxent} shows learning curves for two observation shifts (Rotation, Color-Swap, top row), and two reward shifts (Abrupt $n=1\text{e}6$, Gradual $n=1\text{e}6$, bottom row).
As in previous experiments, on the gradual reward shift all methods perform on-par. 
However, on the Color-Swap and abrupt reward shift experiments, we do observe notable differences. 
We observe significant improvements of PPO+ME over PPO in Color-Swap ($p=6.59\text{e}-7$), and Rotation ($p=8.64\text{e}-6$), however not for experiments where a shift in the reward distribution occurs.
This effect is expected, since essential information about the task is discarded (i.e. color of item changes) in the case of Color-Swap, and the agent needs to re-learn the task, for which a simple form of exploration is already beneficial.
For reward shifts, the agent only needs to update what items to collect after having already learned how to collect them, therefore behaving more randomly does not provide any gains.
A more informative form of exploration as for PP0+ICM yields significant gains over PPO ($p=8.28e-9$), and PPO+max\_ent ($p=1.83\text{e}-8$) for abrupt reward shifts, since the reward model actively steers the policy towards behavior that leads to unexpected reward.
Similarly, PPO+ICM significantly outperforms PPO+ME on Color-Swap ($p=6.62\text{e}-4$), as it enforces exploration of the changed observation space to facilitate re-learning the task at hand.
Combining the two different exploration mechanisms further leads to significant improvements in Color-Swap ($p=4.87\text{e}-2$), however we did not observe the same effect in any other experiment.
On Rotation, PPO+ME, PPO+ICM, and PPO+ICM+ME significantly outperform PPO ($p=8.64\text{e}-6$, $p=4.44\text{e}-9$, $p=7.41\text{e}-8$, respectively). 
In this setup, no significant differences between the different forms of exploration can be observed, indicating that any form of exploration helps.
This is due to the fact that Rotation does not discard essential information about the task, since it preserves color and location of items.
Our results provide compelling evidence that directed exploration mechanisms are more useful for proper reaction to changing environments.

\subsection{Complete Recovery after Domain Shifts}
\begin{figure}[h]
  \centering
  \begin{subfigure}[b]{0.4\textwidth}
    \centering
    \includegraphics[width=\textwidth]{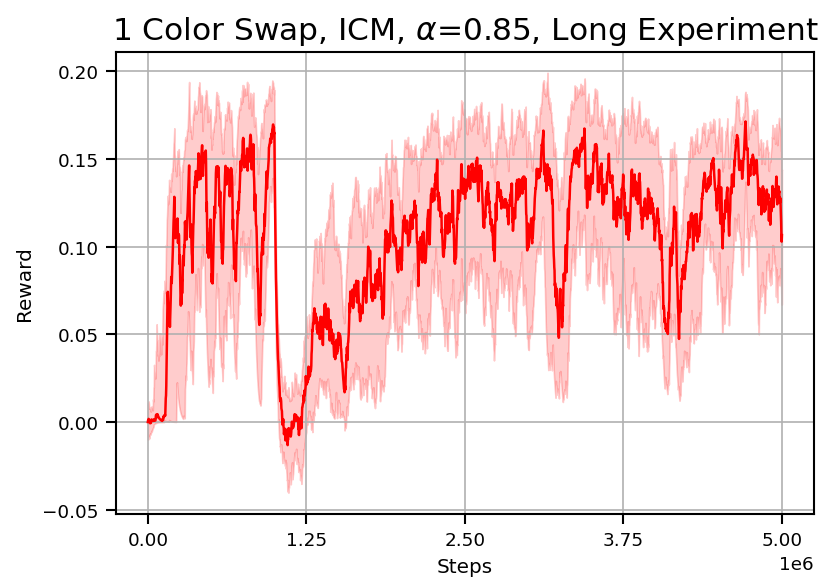}
  \end{subfigure}
    
  \begin{subfigure}[b]{0.1\textwidth}
    \centering
    \includegraphics[width=\textwidth]{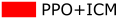}
  \end{subfigure}
  \caption{Experiment for a single abrupt shift after 1M steps (Color-Swap) and another 4M steps without further shifts. PPO+ICM manages to recover all the way, i.e., the achievable reward after a shift is not limited.}
  
  \label{fig:reward_cap}
\end{figure}

Prior experiments (e.g. Figure \ref{fig:non-stationary_dynamics}) may suggest that agents using Reactive Exploration are not capable of recovering to peak performance after a domain shift has occurred.
We show an additional experiment setup in which demonstrates that this is not the case and agents are capable to fully recover given sufficient interaction steps with the environment.
In Figure~\ref{fig:reward_cap} we present results for agents learning for $5\text{e}6$ steps where we apply the Color-Swap shift after $1\text{e}6$ steps. 
The performance of agents converges at first, before the domain shift causes a drop in performance. 
This is followed by recovery and eventually reaching the same performance as before the shift.

\subsection{Generalization to Non-Stationarity}
\label{appendixsub:non-stationary-dynamics}
\begin{figure}[hbt!]
    \centering
    \includegraphics[width=0.8\linewidth]{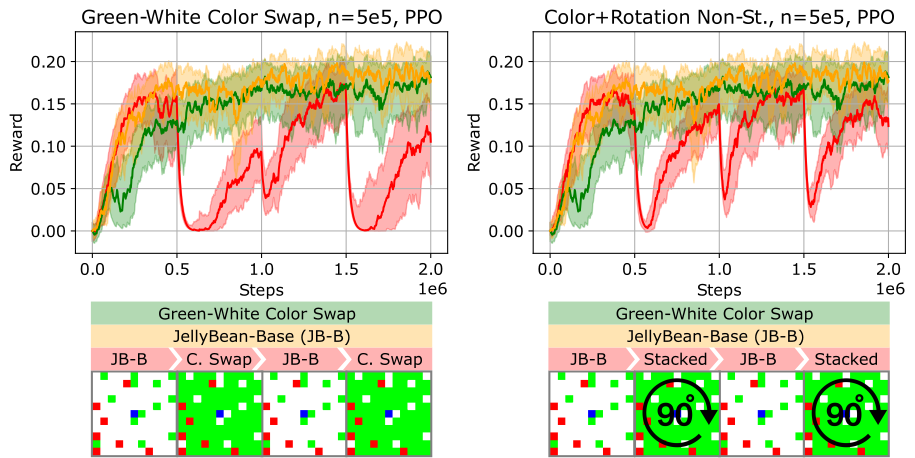}
    \caption{Experiments for non-stationary dynamics via swapping the color of a collectible with the background (GW-Swap, left) versus stacking GW-Swap and rotation (GW-Swap+Rotation, right). PPO agents recover more quickly for the GW-Swap+Rotation. The bottom row visualizes how the non-stationarity occurs. Average rewards every 1000 steps are shown using an IQR and exponential average smoothing across 8 seeds.}
    \label{fig:stacking_non-stationarity}
\end{figure}

As a follow-up experiment, we investigate the generalization behavior of the agent after stacking different domain shifts.
In this regard, we again consider a simple color swap experiment in which the colors green and white (color of collectible which yields positive reward, and background) are exchanged (GW-Swap).
This shift appears every $n=5\text{e}5$ interaction steps.
Figure \ref{fig:stacking_non-stationarity} (left) shows the performance of the stationary versions of PPO and the non-stationary version which encounters GW-Swap.
The PPO agent shows signs of forward transfer and is capable of recovering.
Surprisingly, when combining GW-Swap with an additional transformation (GW-Swap+Rotation), the PPO agent reaches a higher average reward and recovers more quickly (Figure \ref{fig:stacking_non-stationarity}, right).
Also, this particular domain shift appears to facilitate more forward transfer than the single GW-Swap. 
As a reference point, we also provide the results for PPO trained in the stationary regime.

\begin{figure}[hbt!]
    \centering
    \includegraphics[width=0.8\linewidth]{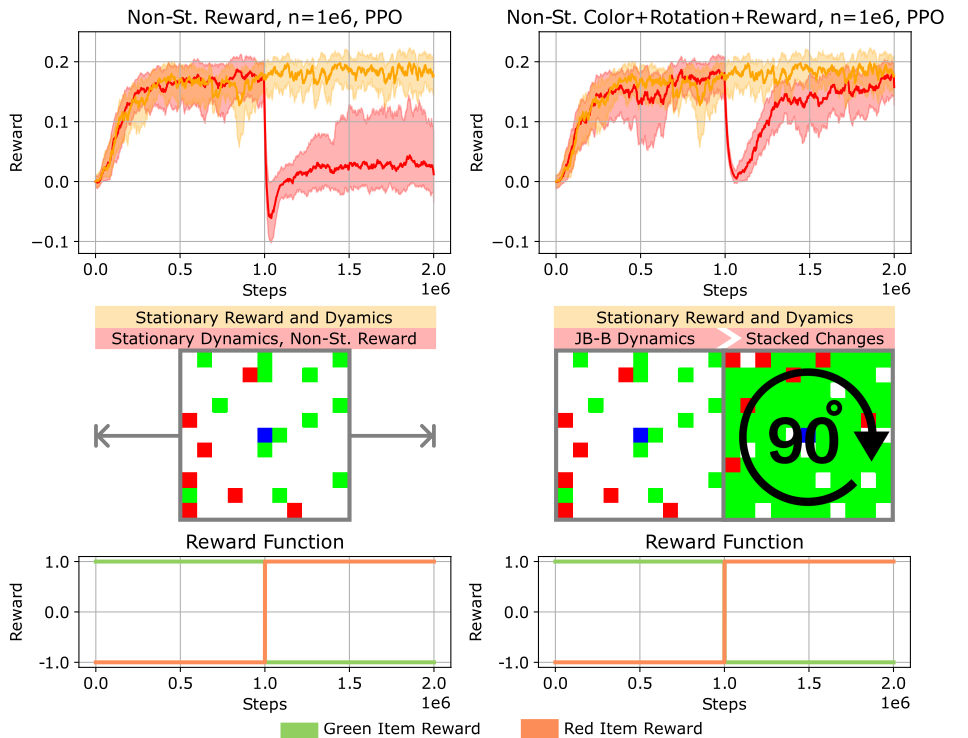}
    \caption{The slow and abrupt non-stationarity experiment from Section~\ref{subsec:ppo_vs_dqn_nonstationarity} (left) versus stacking reward shift with GW-Swap+Rotation (right). The bottom row shows the different kinds of non-stationarity. Average rewards every 1000 steps are shown using an IQR and exponential average smoothing across 8 seeds each. }
    \label{fig:stacking_non-stationarity2}
\end{figure}

We take this line of experiments a step further and additionally induce a shift in the reward function.
This shift occurs after $n=1\text{e}6$ interaction steps with the environment, and inverts the reward function, as explained in Section \ref{subsec:ppo_vs_dqn_nonstationarity}.
Figure \ref{fig:stacking_non-stationarity2}, left, shows the performance for the PPO agent after the shift in the reward function.
When combining this distribution shift with GW-Swap+Rotation, we find that the PPO agent successfully recovers after the induced change.
It appears that certain types of non-stationarities positively influence the ability of the agent to generalize.
We leave a further investigation of this phenomenon to future work.

\section{Experiment Setup}
\label{appendix:experiment-setup}

\subsection{Hyperparameters}
\label{appendixsub:hparams}

\subsubsection{Default DQN and PPO}
\label{appendixsub:hparams-dqn-ppo}
The hyperparameters used in this work are based on prior work on PPO \citep{schulman2017proximal} and DQN \citep{mnih2013playing,mnih2015human}.
These settings are provided as default in the \texttt{Stable-Baselines3}~\citep{stable-baselines3} package.
We experienced that these hyperparameter choices out of the box resulted in reasonable performance for both algorithms, even in the non-stationary case.

\begin{table}[!htb]
    \begin{minipage}{.5\linewidth}
      \centering
        \caption{PPO Hyperparameters}
        \begin{tabular}{ll}
        \multicolumn{1}{l|}{Hyperparameter}                & Value \\ \hline
        \multicolumn{1}{l|}{Learning rate}                 & 3e-4  \\
        \multicolumn{1}{l|}{Rollout collection steps}      & 2048  \\
        \multicolumn{1}{l|}{Batch size}                    & 64    \\
        \multicolumn{1}{l|}{Epochs}                        & 10    \\
        \multicolumn{1}{l|}{$\gamma$}                         & 0.99  \\
        \multicolumn{1}{l|}{GAE $\lambda$}                    & 0.95  \\
        \multicolumn{1}{l|}{Clip range}                    & 0.2   \\
        \multicolumn{1}{l|}{Clip range value function}     & None  \\
        \multicolumn{1}{l|}{Normalize advantage}           & True  \\
        \multicolumn{1}{l|}{Entropy coefficient}           & 0     \\
        \multicolumn{1}{l|}{Value function coefficient}    & 0.5   \\
        \multicolumn{1}{l|}{Maximum gradient clipping}     & 0.5   \\
        \multicolumn{1}{l|}{Resample noise matrix}         & False \\
        \multicolumn{1}{l|}{Limit on Target KL divergence} & None  \\
                                                           &      
        \end{tabular}
    \end{minipage}%
    \begin{minipage}{.5\linewidth}
      \centering
        \caption{DQN Hyperparameters}
        \begin{tabular}{ll}
        \multicolumn{1}{l|}{Hyperparameter}                 & Value \\ \hline
        \multicolumn{1}{l|}{Learning rate}                  & 1e-4  \\
        \multicolumn{1}{l|}{Buffer size}                    & 1e6   \\
        \multicolumn{1}{l|}{Learning starts after}          & 5e4   \\
        \multicolumn{1}{l|}{Batch size}                     & 32    \\
        \multicolumn{1}{l|}{Soft update coefficient $\tau$}                            & 1.0   \\
        \multicolumn{1}{l|}{$\gamma$}                          & 0.99  \\
        \multicolumn{1}{l|}{Update frequency}               & 4     \\
        \multicolumn{1}{l|}{Gradient steps after rollout}   & 1     \\
        \multicolumn{1}{l|}{Target network update interval} & 1e4   \\
        \multicolumn{1}{l|}{$\epsilon$ exploration fraction}   & 0.1   \\
        \multicolumn{1}{l|}{$\epsilon$ initial value}          & 1.0   \\
        \multicolumn{1}{l|}{$\epsilon$ final value}            & 0.05  \\
        \multicolumn{1}{l|}{Maximum gradient clipping}      & 10    \\
                                                            &       \\
                                                            &      
        \end{tabular}
    \end{minipage} 
    \label{tab:hyperparams}
\end{table}

\subsubsection{Mixture of Intrinsic and Extrinsic Rewards}
\label{appendixsubsub:hparams-reactive-exploration}

We use hyperparameters $\alpha$, to weight intrinsic rewards originating from the forward dynamics model, and $\beta$ to weight intrinsic rewards originating from the reward model. 
These parameters control the ratio of intrinsic reward, while $1-\alpha-\beta$ yields the ratio of extrinsic reward received by the agent. 
In addition, it is important to consider the different scales of intrinsic and extrinsic rewards, as the intrinsic rewards generated by the model might be far lower (e.g. 0.01) than the extrinsic rewards of the environment (e.g. 0.15).
This discrepancy can be accounted for in the choice of $\alpha$ and $\beta$.
In certain experiments (i.e. observation shift, Section \ref{subsec:breaking_ppo}) we set $\beta = 0$, as predicting the reward does not provide additional useful information. 

\begin{figure}[hbt!]
        \centering
        \includegraphics[width=1.0\linewidth]{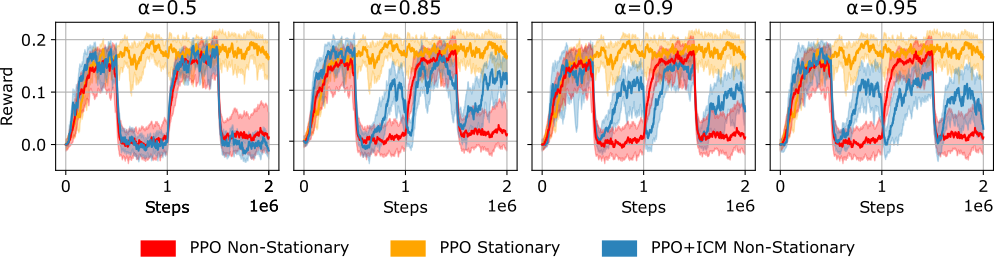}
        \caption{Results for repeating our Color-Swap experiment with added Reactive Exploration. We show different hyperparameter settings for the mixture of intrinsic curiosity reward and extrinsic environment reward over 5 seeds.}
        \label{fig:curiosity-results}
    \end{figure}

In Figure~\ref{fig:curiosity-results} we present four different settings for $\alpha$. 
For the first value of $\alpha=0.5$, the extrinsic reward clearly overshadows the intrinsic reward (roughly 15 times as large), which naturally leads to no Reactive Exploration by the agent, and yields similar results to the default PPO algorithm.
For the additionally considered settings of $\alpha=\{0.85,0.9,0.95\}$ the agent manages to recover from the domain shifts with almost identical behavior. 
The ranges for intrinsic and extrinsic rewards are obtained by a preliminary experiment in which we track the range of the prediction error of the forward dynamics/reward model, and the range of the extrinsic rewards. 
In our particular case, the extrinsic rewards are in an interval of $[0.75;2.65]$ times as large as intrinsic rewards.
Accordingly, we choose the above-mentioned value for $\alpha$. 
A simple and alternative way of ensuring a similar range for intrinsic and extrinsic rewards would be to normalize and scale them, leading to the necessity of a different hyperparameter for mixing intrinsic and extrinsic reward.

\subsection{Model Architectures}
\label{appendixsub:model-architectures}
We apply a custom CNN feature extractor which is small enough to support the (11 $\times$ 11 $\times$ 3) observation space of JBW and additionally contains a linear layer of size 128. 
The CNN feature extractor for both, DQN and PPO, is listed in Table~\ref{table:cnn-feature-extractor}. 
For our CartPole experiments, we use the default MLP policy setting in SB3.

\begin{table}[!h]
\centering
\caption{CNN feature extractor used with Stable-Baselines3 PPO and DQN for JBW.}
\label{table:cnn-feature-extractor}
\begin{tabular}{ l l l }
\hline
\textbf{Layer: Type} & \textbf{Parameters}                                                       & \textbf{Connected to} \\ \hline
0: Conv2D   & in=obs\_channels, out=32, kernel\_size=(3, 3), stride=2  & 1: ReLU      \\ \hdashline
1: ReLU     &                                                                  & 2: Conv2D    \\ \hdashline
2: Conv2D   & in=32, out=64, kernel\_size=(3, 3), stride=2 & 3: ReLU      \\ \hdashline
3: ReLU     &                                                                  & 4: Flatten   \\ \hdashline
4: Flatten  &                                                                  & 5: Linear    \\ \hdashline
5: Linear   & in=n\_flattened, out=128                                                        & 6: ReLU      \\ \hdashline
6: ReLU     &                                                                  &              \\ \hline
\end{tabular}
\end{table}

For the Intrinsic Curiosity Module (ICM~\citealp{pathak2017curiosity}), that is, our observation model, we list the architecture for JBW in Table~\ref{table:icm-architecture-jbw}, and for CartPole in Table~\ref{table:icm-architecture-cartpole}. The corresponding reward model architecture for JBW is listed in Table~\ref{table:reward-model-architecture-jbw}, and that for CartPole in Table~\ref{table:reward-model-architecture-cartpole}.

\begin{table}[!h]
\centering
\caption{ICM observation model architecture for JBW consisting of inverse and forward model which share an observation encoder.}
\label{table:icm-architecture-jbw}
\begin{tabular}{ l l l }
\hline
\textbf{Layer: Type}           & \textbf{Parameters}                                                                        & \textbf{Connected to}    \\ \hline
enc\_0: Conv2D        & in=obs\_channels, out=32, kernel\_size=(3, 3), stride=2 & enc\_1: ReLU    \\ 
\hdashline

enc\_1: ReLU          &                                                                                   & enc\_2: Conv2D  \\ \hdashline
enc\_2: Conv2D        & in=32, out=64, kernel\_size=(3, 3), stride=2                  & enc\_3: ReLU    \\ \hdashline
enc\_3: ReLU          &                                                                                   & enc\_4: Flatten \\ \hdashline
enc\_4: Flatten       &                                                                                   & enc\_5: Linear  \\ \hdashline
enc\_5: Linear        & out=128                                                                         & enc\_6: ReLU    \\ \hdashline
enc\_6: ReLU          &                                                                                   &                 \\ \hdashline
forw\_0:  Concatenate & input action, input observation                       & forw\_1: Linear \\ \hdashline
forw\_1: Linear       & action\_len + observation\_latent\_features, hdashline                           & forw\_2: ReLU   \\ \hdashline
forw\_2: ReLU         &                                                                                   & forw\_3: Linear \\ \hdashline
forw\_3: Linear       & in=128, out=128                                                                   & forw\_4: ReLU   \\ \hdashline
forw\_4: ReLU         &                                                                                   & forw\_5: Linear \\ \hdashline
forw\_5: Linear       & in=128, out=128                                                                       &                 \\ \hdashline
inv\_0: Linear        & in=2 $\cdot$ observation\_latent\_features, out=128                                             & inv\_1: ReLU    \\ \hdashline
inv\_1: ReLU          &                                                                                   & inv\_2: Linear  \\ \hdashline
inv\_2: Linear        & in=128, out=128                                                                   & inv\_3: ReLU    \\ \hdashline
inv\_3: ReLU          &                                                                                   & inv\_4: Linear  \\ \hdashline
inv\_4: Linear        & in=128, out=action\_len                                              &                 \\ \hline
\end{tabular}
\end{table}

\begin{table}[!h]
\centering
\caption{ICM observation model encoder for CartPole, reusing the forward and inverse models described in Table~\ref{table:icm-architecture-jbw}.}
\label{table:icm-architecture-cartpole}
\begin{tabular}{l l l}
\hline
\textbf{Layer: Type}    & \textbf{Parameters}             & \textbf{Connected to}    \\ \hline
enc\_0: Linear & in=observation\_len, out=128 & enc\_1: ReLU    \\ \hdashline
enc\_1: ReLU   &                        & enc\_2: Conv2D  \\ \hdashline
enc\_2: Linear & in=128, out=128               & enc\_3: ReLU    \\ \hdashline
enc\_3: ReLU   &                        & enc\_4: Flatten \\ \hdashline
enc\_4: Linear & in=128, out=128               & enc\_5: Linear  \\ \hline
\end{tabular}
\end{table}

\begin{table}[h]
\centering
\caption{Reward model architecture for JBW.}
\label{table:reward-model-architecture-jbw}
\begin{tabular}{l l l}
\hline
\textbf{Layer: Type} & \textbf{Parameters}                                              & \textbf{Connected to} \\ \hline
0: Conv2D   & in=obs\_channels, out=32, kernel\_size=(3, 3), stride=2 & 1: ReLU      \\ \hdashline
1: ReLU     &                                                         & 2: Conv2D    \\ \hdashline
2: Conv2D   & in=32, out=64, kernel\_size=3, stride=2                 & 3: ReLU      \\ \hdashline
3: ReLU     &                                                         & 4: Flatten   \\ \hdashline
4: Flatten  &                                                         & 5: Concat    \\ \hdashline
5: Concat   & action\_len, observation\_latent\_features       & 6: Linear    \\ \hdashline
6: Linear   & in=action\_len + n\_flatten, out=128       & 7: ReLU      \\ \hdashline
7: ReLU     &                                                         & 8: Linear    \\ \hdashline
8: Linear   & in=128, out=128                                         & 9: ReLU      \\ \hdashline
9: ReLU     &                                                         & 10: Linear   \\ \hdashline
10: Linear  & in=128, out=1                                           &              \\ \hline
\end{tabular}
\end{table}

\begin{table}[h]
\centering
\caption{Reward model architecture for CartPole.}
\label{table:reward-model-architecture-cartpole}
\begin{tabular}{l l l}
\hline
\textbf{Layer: Type} & \textbf{Parameters}                         & \textbf{Connected to} \\ \hline
0: Concat   & action, observation                      & 1: Linear    \\ \hdashline
1: Linear   & in=action\_len+observation\_len, out=128 & 2: ReLU      \\ \hdashline
2: ReLU     &                                    & 3: Linear    \\ \hdashline
3: Linear   & in=128, out=128                    & 4: ReLU      \\ \hdashline
4: ReLU     &                                    & 5: Linear    \\ \hdashline
5: Linear   & in=128, out=128                    & 6: ReLU      \\ \hdashline
6: ReLU     &                                    & 7: Linear    \\ \hdashline
7: Linear   & in=128, out=1                      &              \\ \hline
\end{tabular}
\end{table}

\end{document}